\documentclass[11pt]{article}
	
	\newcommand{\blind}{0}
	
    \makeatletter
    \renewcommand\section{\@startsection {section}{1}{\z@}%
                                       {-3.5ex \@plus -1ex \@minus -.2ex}%
                                       {2.3ex \@plus.2ex}%
                                       {\normalfont\fontfamily{phv}\fontsize{16}{19}\bfseries}}
    \renewcommand\subsection{\@startsection{subsection}{2}{\z@}%
                                         {-3.25ex\@plus -1ex \@minus -.2ex}%
                                         {1.5ex \@plus .2ex}%
                                         {\normalfont\fontfamily{phv}\fontsize{14}{17}\bfseries}}
    \renewcommand\subsubsection{\@startsection{subsubsection}{3}{\z@}%
                                        {-3.25ex\@plus -1ex \@minus -.2ex}%
                                         {1.5ex \@plus .2ex}%
                                         {\normalfont\normalsize\fontfamily{phv}\fontsize{14}{17}\selectfont}}
    \makeatother
	
	\usepackage{amsmath}
	\usepackage{graphicx}
	\usepackage{enumerate}
	\usepackage{natbib} 
	\usepackage{url} 
    \usepackage{xcolor}
	
    \usepackage{booktabs}
    \usepackage{amssymb}   
    \usepackage{multirow}
    \usepackage{subcaption}
    \usepackage{comment}
    \usepackage{algorithm}
    \usepackage{algpseudocode}
    \usepackage[hidelinks]{hyperref}
    \usepackage{amsthm}
    \theoremstyle{definition}
    
    \usepackage{makecell}
    \usepackage{soul}
    \usepackage{longtable}
	
\begin{document}
		
		\def\spacingset#1{\renewcommand{\baselinestretch}%
			{#1}\small\normalsize} \spacingset{1}
		
		\if0\blind
		{
			\title{\bf Federated Physics-Informed Thermal Field Prediction Across Unseen Geometries in Distributed Additive Manufacturing}
			\author{Amirul Islam Saimon and Zhenyu James Kong$^\ast$ (email: zkong@vt.edu) \\[7pt]
			Grado Department of Industrial and Systems Engineering, Virginia Tech,\\ 
            Blacksburg, USA}
			\date{}
			\maketitle
		} \fi
		
		\if1\blind
		{

            \title{\bf Federated Physics-Informed Thermal Field Prediction Across Unseen Geometries in Distributed Additive Manufacturing}
			\author{Author information is purposely removed for double-blind review}
			
\bigskip
			\bigskip
			\bigskip
			\begin{center}
				{\LARGE\bf Federated Physics-Informed Thermal Field Prediction Across Unseen Geometries in Distributed Additive Manufacturing}
			\end{center}
			\medskip
		} \fi
		\bigskip
		
	\begin{abstract}
        In additive manufacturing (AM), each change in part geometry requires a new thermal simulation, which makes design evaluation over many candidates costly. Geometry-aware models reduce this cost by predicting thermal fields on unseen parts with a single learned model, but they assume that diverse geometries are available centrally. This assumption is restrictive when each manufacturing site holds only a limited set of proprietary part designs that it cannot share. We propose Fed-PIGANO, a federated framework that trains physics-informed geometry-aware neural operator (PI-GANO) models at individual sites and aggregates their geometry encoders and prediction networks without sharing part geometries. We evaluate the framework on a 2D steady-state thermal benchmark for metal AM, with geometries from five polygonal families distributed across three sites under four scenarios that range from identical to strongly different family compositions. Over seven replications, Fed-PIGANO achieves mean relative \(\ell_2\) errors of 1.83\%–2.04\% across all 12 scenario–site cases, compared with up to 20.46\% for Local-only training, and consistently yields lower PDE residuals. Its prediction accuracy is comparable to that of a centralized model trained on the pooled geometries of all sites. These results demonstrate that federated learning lets manufacturing sites benefit from each other's geometries without sharing them, enabling rapid thermal prediction on new part designs for tasks such as shape optimization, inverse design, and uncertainty quantification.
	\end{abstract}
			
	\noindent%
	{\it Keywords:} Distributed manufacturing; Physics-informed learning; Federated learning; Geometry-aware learning; Thermal field prediction; Additive manufacturing.

	\spacingset{1.5} 

\section{Introduction}
\label{sec:Fed_introduction}
Many physical phenomena in engineering, including heat transfer, fluid flow, and structural deformation, are described by systems of partial differential equations (PDEs). Numerical methods such as the finite element method (FEM) approximate their solutions by discretizing the geometric domain into a mesh \citep{Johnson2009_numerical_PDE_by_FEM_book_ref}. For systems with multiscale and multiphysics behavior, however, such solvers become computationally demanding \citep{Smith2016_thermal_importance_AM_Supp_ref, Saimon_Zaman_2026_WMLE_LAMDA}. This cost is particularly limiting in many-query tasks such as design optimization, which may require a large number of simulations \citep{Gao_Sun__Wang2021_PhyGeoNet_CNN_irregular_domain,
Liao2023_Thermal_AM_PINN_ref}.

The design phase of thermal-based metal additive manufacturing (AM) is a representative many-query problem, where candidate part geometries are repeatedly evaluated and revised \citep{Saimon2026_DL4AM, Almasri2024_DfAM_motivation}. In these processes, a localized heat source melts material that then solidifies, building up the part layer by layer \citep{Rongxuan_Kong2023_LPBF_control}. The source may be stationary, as in spot melting, where a laser exposes a single location \citep{Guaglione_Demir2021_Single_Point_exposure_reference_spot_melting, Rongxuan__Kong2022_Spot_melting_Reference, Ruiyuan_Zhang2025automatic_spot_melting_ref}. It may also move across a powder bed, as in laser powder bed fusion (L-PBF) \citep{Rongxuan_Kong2023_LPBF_control}. In both cases, the heat diffusion PDE governs the thermal behavior over the part's specific geometry. It couples material properties (e.g., thermal conductivity, density, specific heat), process parameters (e.g., heat source power, beam size, scan speed), and boundary interactions with the surroundings \citep{Faegh__Rao_Jin2025_PINN_for_PSP_in_AM_review}. The resulting thermal history of the part is central to its process--structure--property (PSP) relationship. It drives microstructure evolution, which in turn determines the material's mechanical properties and, ultimately, the manufactured part's quality \citep{Smith2016_thermal_importance_AM_Supp_ref, Deshmukh_Rao2024_thermal_on_quality_Supp_Ref}. A representative thermal-based PSP framework for a fixed geometry is presented in our earlier work \citep[see Fig.~1 of][]{Saimon2026_GeoPIPN}.

Under fixed process parameters, any change to the part shape changes the domain boundary and therefore the thermal field. Such changes are common during the design phase of AM, for example to improve performance, reduce material usage, or meet new requirements \citep{Saimon2026_DL4AM}. Design tasks such as shape optimization, inverse design, uncertainty quantification, and design exploration follow an iterative loop around these changes. In iteration $i$, the thermal field $T(x,y;\Omega^{(i)})$ of a candidate geometry $\Omega^{(i)}$ is evaluated, and a performance metric is computed from it (e.g., peak temperature, cooling rate, or a microstructure-related quality index). The geometry is then updated to $\Omega^{(i+1)}$, for example by a gradient-based or heuristic search step. Each update changes the thermal field, so the structure and property predictions computed from the previous field no longer apply. In the conventional workflow, the new geometry must then be meshed and solved with FEM again. As iterations accumulate, this repeated cost substantially lengthens the part development cycle \citep{Saimon2026_DL4AM}.

Physics-informed neural networks (PINNs) \citep{Raissi_Perdikaris_Karniadakis2019_PINN_original} incorporate the governing PDE and boundary conditions directly into model training, which reduces the dependence on labeled thermal solutions. Conventional PINN formulations for AM, however, consider a fixed geometric domain, so a change in part geometry typically requires retraining the model \citep{Saimon2026_GeoPIPN, Faegh__Rao_Jin2025_PINN_for_PSP_in_AM_review}. Geometry-aware physics-informed models remove this limitation by taking the domain geometry as an input to the prediction model. Once trained on a set of known geometries $\{\Omega^{(1)},\dots,\Omega^{(n)}\}$, such a model with trainable parameters $\boldsymbol{\theta}$ predicts the thermal field $T_{\boldsymbol{\theta}}(x,y;\Omega^{\mathrm{new}})$ on an unseen geometry $\Omega^{\mathrm{new}}$ without retraining or a new FEM solve \citep{Zhong_Meidani2025_PI-GANO_both_Geometry_parameters, Saimon2026_GeoPIPN}. Each design iteration then needs only a model evaluation, and the microstructure, property, and quality predictions that build on the thermal field are obtained faster as well \citep{Saimon2026_GeoPIPN}.

\begin{figure}[htbp]
  \centering
    \includegraphics[width=0.75\linewidth]{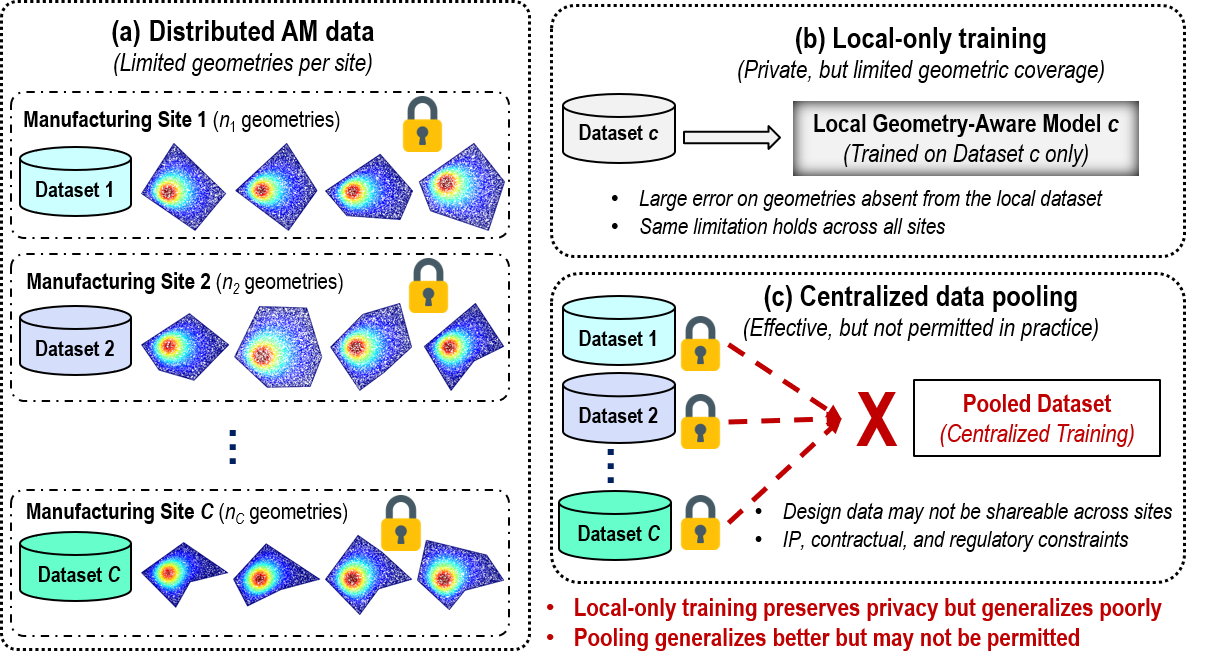}
    \caption{Distributed-data challenge in AM. (a) Each of the $C$ sites holds a limited set of $n_{c}$ part geometries. (b) Local-only training keeps data private but covers few geometries. (c) Pooling broadens coverage but may be restricted by data-sharing constraints.}
    \label{fig:motivation_federated_geometry_generalization}
\end{figure}

The ability of such models to generalize to unseen part shapes depends on the geometric diversity of the known geometries used for training \citep{Zhong_Meidani2025_PI-GANO_both_Geometry_parameters, Saimon2026_GeoPIPN}. In distributed AM, however, design and production take place independently at multiple manufacturing sites, and each site accumulates its own geometry dataset, shaped by the parts it has produced (Fig.~\ref{fig:motivation_federated_geometry_generalization}a). A site $c$ may therefore hold only $n_{c}$ geometries, which cover only part of the geometric variation across the manufacturing network.

This distributed setting introduces two practical constraints for training geometry-aware models. First, a model trained only on a site's local dataset keeps its geometries private but sees limited geometric diversity (Fig.~\ref{fig:motivation_federated_geometry_generalization}b), so it may predict poorly on geometries that are rare or absent locally. Second, pooling the datasets of multiple sites would broaden geometric coverage (Fig.~\ref{fig:motivation_federated_geometry_generalization}c), but such data sharing may be restricted by proprietary design information, intellectual property considerations, contractual agreements, or regulatory requirements. Each site could instead add synthetic geometries, but these may not reflect the part designs produced at other sites.

These constraints motivate the use of federated learning (FL) \citep{McMahan2017_Communication_FedAvg, Kairouz_McMahan2019_advances_and_open_problems_FL, Li__Smith2020_FL-challenges_methods_and_future_directions}, which enables multiple manufacturing sites to collaboratively train a shared model while keeping their part geometries local. Each site trains the model using its own local data and communicates only model parameters to a coordinating server, where the updates are aggregated into a shared global model over successive communication rounds. In our setting, all sites share the same metal AM thermal process, with the same steady-state heat conduction equation, Gaussian laser source, and material properties. Under this setting, site-to-site differences arise from differences in the local distributions of part geometries rather than from differences in the underlying thermal physics or process conditions (Fig.~\ref{fig:motivation_federated_geometry_generalization}a). Federated training must therefore aggregate both the geometry encoder and the prediction network of the local models. The resulting global model learns from the geometry families of all sites and can predict on unseen geometries beyond any single site's training set, without pooling the geometries. 

In this paper, we make the following contributions:
\begin{itemize}
    \item We formulate federated thermal-field prediction across unseen AM part geometries, where manufacturing sites collaboratively train a geometry-aware model without pooling proprietary part designs.
    \item We propose Fed-PIGANO, a federated physics-informed geometry-aware framework that adopts PI-GANO as the local model at each site, trains it with a physics-informed loss without labeled temperature data, and aggregates its geometry encoder and prediction network across sites.
    \item We evaluate Fed-PIGANO on 2D metal AM geometries from five families, held by three sites under four scenarios with increasingly different family compositions, over seven replications. Its mean relative error on unseen geometries stays between 1.8\% and 2.0\% at every site, whereas that of training at each site alone (Local-only) reaches 20.5\%. Its accuracy is also comparable to that of a centralized model trained on the pooled geometries.
\end{itemize}

The remainder of this paper is organized as follows. Section~\ref{sec:Fed_related_work} reviews three lines of work that converge in our approach. Section~\ref{sec:Fed_methodology} formulates the problem and presents Fed-PIGANO. Section~\ref{sec:Fed_experimental_setup} describes the experimental setup, followed by the results and discussion in Section~\ref{sec:Fed_results_discussion}. Finally, Section~\ref{sec:fed_conclusion_future_work} concludes the paper and outlines directions for future extension.

\section{Related Work}
\label{sec:Fed_related_work}
This section reviews physics-informed thermal modeling in AM (Section~\ref{sec:PINN_for_AM_related_work}), geometry-aware physics-informed learning (Section~\ref{sec:geomety_aware_PINN_related_work}), and federated learning in manufacturing and scientific machine learning (Section~\ref{sec:FL_in_mfg_and_sml_related_work}). Section~\ref{sec:research_gap_related_work} then summarizes the research gap at their intersection.

\subsection{\emph{Physics-Informed Thermal Modeling in Additive Manufacturing}} \label{sec:PINN_for_AM_related_work}

Physics-informed learning \citep{Raissi_Perdikaris_Karniadakis2019_PINN_original, Gnanasambandam2023_Stan_Raghav} has been increasingly explored for thermal modeling in AM \citep{Zhu2021_Thermal_AM_PINN_ref, Xie20223_Thermal_AM_PINN_ref, Liao2023_Thermal_AM_PINN_ref, Gnanasambandam2026_CoPINN_AFSD}. A broader review of PINN applications in AM is provided by \citet{Faegh__Rao_Jin2025_PINN_for_PSP_in_AM_review}. These AM studies primarily consider a fixed part geometry during training and inference, referred to as the \textit{single-geometry PINN} setting by \citet[Fig.~2, \textit{top}]{Saimon2026_GeoPIPN}. In this setting, the model predicts at unseen coordinates within the same geometric domain, but a change in part geometry requires retraining the model. This limits its use in design tasks that repeatedly evaluate different geometries \citep{Kashefi_Mukerji2022_PIPN_original, Saimon2026_DL4AM, Zhong_Meidani2025_PI-GANO_both_Geometry_parameters}. Geometry-aware physics-informed models, reviewed in Section~\ref{sec:geomety_aware_PINN_related_work}, address this limitation by learning across multiple geometries.

\subsection{\emph{Geometry-Aware Physics-Informed Learning}} \label{sec:geomety_aware_PINN_related_work}

PIPN \citep{Kashefi_Mukerji2022_PIPN_original} introduced geometry-aware physics-informed learning by combining PINNs \citep{Raissi_Perdikaris_Karniadakis2019_PINN_original} with the point-cloud-based PointNet architecture \citep{Qi_Su__Guibas2017_PointNet}. PIPN and its extensions have been applied to fluid flow, structural mechanics, porous-media transport, aerodynamics, vascular flow, crack propagation, and acoustic--structure interaction \citep{Kashefi_Mukerji2022_PIPN_original, Kashefi_Guibas_Mukerji2023_On_how_many_geom_PIPN, Zhong_Meidani2025_PI-GANO_both_Geometry_parameters, Kashefi2023_PIPN_application_porous_media_ref, Lin_You__Hu2024_Data_driven_PIPN_bluff_bodies_ref, Kang2024_Ref_application_of_PIPN, Tu__Qi2022_PINN_PointNet_Crack_Propagation, Wang2024_Mpipn_PIPn_application_to_acoustic_structure_systems}. More recently, PI-KAN-PointNet \citep{Kashefi_2026_PI-KAN-PointNet} replaced the shared MLPs of the PIPN architecture with Kolmogorov--Arnold Networks (KANs) \citep{Liu__Tegmark2025_KAN_original_paper}. GeoPIPN \citep{Saimon2026_GeoPIPN} extends geometry-aware physics-informed prediction to thermal fields on unseen AM part geometries. PI-GANO \citep{Zhong_Meidani2025_PI-GANO_both_Geometry_parameters}, in contrast, is a general physics-informed geometry-aware neural operator that handles variations in both geometry and PDE parameters. Generalization in scientific machine learning may thus concern the geometric domain, the PDE parameters, or both. This work focuses on \textit{geometry generalization} with fixed material and process parameters, and we adopt the geometry-only variant of PI-GANO as the local model at each site (Section~\ref{subsec:pigano_arch}). All of these approaches, however, train on a centralized dataset and assume that all training geometries are accessible in one place. This assumption does not hold when the geometries are held by different AM sites.

\subsection{\emph{Federated Learning in Manufacturing and Scientific Machine Learning}} \label{sec:FL_in_mfg_and_sml_related_work}

FL has been explored in manufacturing, particularly in AM, to enable collaborative modeling while keeping the data at local sites. \citet{Mehta__Shao2024federated_dimension_prediction_AM} applied FL to dimensional accuracy prediction and part qualification across AM factories. For metal L-PBF, \citet{Mehta_Shao2022federated_semantic_segmentation_defect_detection_AM} considered federated defect detection, while \citet{Lei__Guo2024bridging_data_gaps_heat_emission_prediction_LPBF} investigated federated heat-emission prediction across small and medium-sized enterprises. Collaborative learning has also been investigated for process-parameter recommendation across networked 3D printers \citep{Guo__Kontar_Okwudire2026_collaborative_process_param_recommender_system}. These studies show that collaborative learning can benefit distributed manufacturing, but they do not consider physics-informed thermal field prediction across varying AM part geometries. 

In the broader scientific machine learning literature, \citet{Zhang__LuLu2025federated_SciML_functions_PDE_data_heterogeneity} presented a systematic study of federated learning for function approximation and PDE solving. They introduced FedAvg-Adam, a training strategy that replaces the stochastic gradient descent of FedAvg \citep{McMahan2017_Communication_FedAvg} with Adam for local updates. In their federated PINN formulation, the collocation points of a single square domain are divided among clients. All clients thus solve the same PDE on the \textit{same} domain, and the samples distributed across clients are \textit{spatial coordinates rather than different domains}. This formulation therefore does not require a geometry encoder to represent different domain shapes across clients, and the study does not address AM thermal problems.

\subsection{\emph{Research Gap}}
\label{sec:research_gap_related_work}

Table~\ref{tab:Fed_related_work_and_research_gap} summarizes the three lines of work reviewed above. Physics-informed thermal models for AM (Section~\ref{sec:PINN_for_AM_related_work}) mostly consider a single fixed geometry. Geometry-aware physics-informed models (Section~\ref{sec:geomety_aware_PINN_related_work}) predict across varying domains but assume that all training geometries are available in one place. Federated learning studies in manufacturing and scientific machine learning (Section~\ref{sec:FL_in_mfg_and_sml_related_work}) keep data local but do not consider varying geometric domains across clients. To our knowledge, no existing work combines physics-informed learning, geometry generalization, and federated training for AM thermal-field prediction. This work addresses this gap with Fed-PIGANO, which trains a geometry-aware physics-informed model across AM sites without pooling their part geometries.

\begin{table}[htbp]
    \centering
    \footnotesize
    \caption{Comparison of related studies and the research gap addressed in this work.}
    \label{tab:Fed_related_work_and_research_gap}
    \renewcommand{\arraystretch}{0.90}
    \begin{tabular*}{\textwidth}{@{\extracolsep{\fill}}lcccc@{}}
        \toprule
        Study & \makecell{Physics-\\informed} & \makecell{Geometry-\\aware} & \makecell{Federated\\learning} & \makecell{AM thermal\\focus} \\
        \midrule
        PIPN \citep{Kashefi_Mukerji2022_PIPN_original} & \checkmark & \checkmark & $\times$ & $\times$ \\
        PI-KAN-PointNet \citep{Kashefi_2026_PI-KAN-PointNet} & \checkmark & \checkmark & $\times$ & $\times$ \\
        PI-GANO \citep{Zhong_Meidani2025_PI-GANO_both_Geometry_parameters} & \checkmark & \checkmark & $\times$ & $\times$ \\
        coPINN \citep{Gnanasambandam2026_CoPINN_AFSD} & \checkmark & $\times$ & $\times$ & \checkmark \\
        GeoPIPN \citep{Saimon2026_GeoPIPN} & \checkmark & \checkmark & $\times$ & \checkmark \\
        FedAvg-Adam \citep{Zhang__LuLu2025federated_SciML_functions_PDE_data_heterogeneity} & \checkmark & $\times$ & \checkmark & $\times$ \\
        \midrule
        This work & \checkmark & \checkmark & \checkmark & \checkmark \\
        \bottomrule
    \end{tabular*}
\end{table}

\section{Methodology}
\label{sec:Fed_methodology}
This section first formulates thermal prediction on unseen geometries across AM sites (Section~\ref{sec:Fed_problem_statement}) and the thermal model that all sites share (Section~\ref{sec:Fed_problem}). It then presents the Fed-PIGANO framework (Section~\ref{sec:fedavg_adam}), followed by its local geometry-aware model (Section~\ref{subsec:pigano_arch}) and physics-informed loss (Section~\ref{sec:Fed_loss}). In this paper, we consider 2D AM part geometries, so all equations in this section are written in the in-plane coordinates $(x,y)$.

\subsection{\emph{Problem Formulation for Cross-Site Thermal Prediction on Unseen Geometries}}
\label{sec:Fed_problem_statement}

We consider $C$ clients, indexed by $c \in \{1,\ldots,C\}$, each representing an AM site that designs and builds parts independently. All clients share the thermal problem of Section~\ref{sec:Fed_problem}. They differ only in the part geometries they hold, which reflect each site's own design history and may cover only a subset of the geometric variation across the network. In our experiments, this site-to-site difference takes the form of different geometry-family compositions (Section~\ref{sec:client_distributions}). Client $c$ holds a training set $\mathcal{D}_{c}^{\mathrm{tr}}$ of $n_{c}$ geometries, a validation set $\mathcal{D}_{c}^{\mathrm{val}}$, and a test set $\mathcal{D}_{c}^{\mathrm{te}}$. Each geometry $g$ is a domain $\Omega^{g} \subset \mathbb{R}^{2}$ that represents a 2D AM part (Section~\ref{subsec:thermal_model}). It is discretized by a point set $\mathbf{X}^{g} = \mathbf{X}_{b}^{g} \cup \mathbf{X}_{\mathrm{int}}^{g}$, with $n_{b}$ boundary points $\mathbf{X}_{b}^{g} \subset \partial\Omega^{g}$ and $n_{\mathrm{int}}$ interior points $\mathbf{X}_{\mathrm{int}}^{g} \subset \Omega^{g}$, so that $N_{\mathrm{pts}} = n_{b} + n_{\mathrm{int}}$ is the same for all geometries. The boundary point set $\mathbf{X}_{b}^{g}$ describes the shape of the geometry, and the full point set $\mathbf{X}^{g}$ provides the collocation points for training. All coordinates are in the nondimensional form of Section~\ref{subsec:nondim}.

The goal is a single geometry-aware model $T_{\boldsymbol{\theta}}(x,y;\mathbf{X}_{b})$ that predicts the temperature at any point $(x,y)$ of a geometry $\Omega$ described by its boundary point set $\mathbf{X}_{b}$, including new candidate geometries that no client has seen, such as those that arise when part designs are revised (Section~\ref{sec:Fed_introduction}). The geometries remain at their clients, and only model parameters are exchanged with a central server. The federated objective is
\begin{equation}
  \min_{\boldsymbol{\theta}} \; \mathcal{J}(\boldsymbol{\theta})
  = \sum_{c=1}^{C} w_{c}\, \mathcal{L}\!\left(\boldsymbol{\theta};\,\mathcal{D}_{c}^{\mathrm{tr}}\right),
  \qquad
  w_{c} = \frac{n_{c}}{\sum_{c'=1}^{C} n_{c'}}
  \label{eq:global_obj}
\end{equation}
where $\mathcal{L}$ is the physics-informed loss of Section~\ref{sec:Fed_loss}, and the weights $w_{c}$ follow FedAvg~\citep{McMahan2017_Communication_FedAvg}. When all clients hold the same number of geometries, as in our experiments, $w_{c} = 1/C$.

\subsection{\emph{AM Thermal Model}}\label{sec:Fed_problem}
This subsection defines the steady-state heat conduction problem that all clients share and the nondimensional form used throughout the rest of the paper.

\subsubsection{\emph{Steady-State Heat Conduction}}
\label{subsec:thermal_model}
We consider steady-state heat conduction on a geometry $\Omega$, with the superscript $g$ dropped. The domain represents a thin horizontal cross-section of the part, i.e., an effective deposited layer of thickness $h$, heated by a Gaussian laser source. The material properties, laser source, and boundary condition are the same for every geometry and every client, so the part geometry is the only quantity that changes the thermal field. The governing equation is
\begin{equation}
    -k \left( \frac{\partial^{2} T(x,y)}{\partial x^{2}} + \frac{\partial^{2} T(x,y)}{\partial y^{2}} \right) = q(x,y),
    \qquad (x,y) \in \Omega
\label{eq:gov_dim}
\end{equation}
subject to the Dirichlet boundary condition
\begin{equation}
    T(x,y) = T_{a}, \qquad (x,y) \in \partial \Omega
\label{eq:bc_dim}
\end{equation}
where $k$ is the thermal conductivity, $T$ is the temperature, $T_{a}$ is the ambient temperature, and $q$ is the volumetric heat source. The laser source is modeled as a circular Gaussian beam with absorbed power $Q_{\mathrm{total}} = \alpha P$, where $P$ is the laser power and $\alpha$ is the absorptivity:
\begin{equation}
    q(x,y) = \frac{2 Q_{\mathrm{total}}}{\pi r_{b}^{2} h}
    \exp\!\left( -\frac{2\left[ (x - x_{L})^{2} + (y - y_{L})^{2} \right]}{r_{b}^{2}} \right)
\label{eq:source_dim}
\end{equation}
where $r_{b}$ is the beam radius and $(x_{L}, y_{L})$ is the beam center. Its prefactor makes the integral of the source over the layer, $h \int_{\mathbb{R}^{2}} q(x,y) \,\mathrm{d}x\,\mathrm{d}y$, equal to $Q_{\mathrm{total}}$, so the absorbed power is distributed as a Gaussian in the plane and uniformly through the layer thickness.

The source is stationary, as in spot melting, where the laser exposes a single location \citep{Guaglione_Demir2021_Single_Point_exposure_reference_spot_melting, Rongxuan__Kong2022_Spot_melting_Reference, Ruiyuan_Zhang2025automatic_spot_melting_ref}. We assume that the exposure lasts long enough for the temperature to reach a steady state, and solve Eq.~\eqref{eq:gov_dim} as a boundary value problem with the source held at a fixed position $(x_{L}, y_{L})$. The resulting field approximates the temperature during one exposure. Table~\ref{tab:case_params} lists the source position together with all material and process parameters.

\begin{table}[htbp]
    \centering
    \footnotesize
    \caption{Material and process parameters of the thermal model.}
    \label{tab:case_params}
    \renewcommand{\arraystretch}{0.90}
    \begin{tabular*}{\textwidth}{@{\extracolsep{\fill}}llc@{\hspace{1.5em}}llc@{}}
        \toprule
        Parameter & Symbol & Value & Parameter & Symbol & Value \\
        \midrule
        Thermal conductivity & $k$ & $9.8$~W\,m$^{-1}$\,K$^{-1}$ & Laser power & $P$ & $375$~W \\
        Ambient temperature & $T_{a}$ & $300$~K & Absorptivity & $\alpha$ & $0.4$ \\
        Effective layer thickness & $h$ & $2.5$~mm & Absorbed power & $Q_{\mathrm{total}}=\alpha P$ & $150$~W \\
        Beam radius & $r_{b}$ & $2.5$~mm & Source center & $(x_{L}, y_{L})$ & $(-1.6\,r_{b},\,0)$ \\
        \bottomrule
    \end{tabular*}
\end{table}

\subsubsection{\emph{Nondimensionalization}}
\label{subsec:nondim}

The parameters in Table~\ref{tab:case_params} span several orders of magnitude. We therefore nondimensionalize the problem, which combines them into a single source amplitude and brings the network inputs, the output, and the PDE terms to order one. This matches the scale assumed by standard weight initialization and keeps the boundary and PDE loss terms comparable in magnitude, which improves the conditioning of training \citep{Wang__Perdikaris2023_Expert_guide_to_training_PINNs}. We use the beam radius $r_{b}$ as the length scale and the ambient temperature $T_{a}$ as the temperature scale:
\begin{equation}
    x = r_{b}\,x^{*}, \qquad y = r_{b}\,y^{*}, \qquad T = T_{a}\,T^{*}
\label{eq:scales}
\end{equation}
With these scales, $\partial^{2}/\partial x^{2} = r_{b}^{-2}\,\partial^{2}/\partial x^{*2}$ and $\partial^{2}/\partial y^{2} = r_{b}^{-2}\,\partial^{2}/\partial y^{*2}$, and the exponent of Eq.~\eqref{eq:source_dim} becomes $-2\left[ (x^{*} - x_{L}^{*})^{2} + (y^{*} - y_{L}^{*})^{2} \right]$ with $x_{L}^{*} = x_{L}/r_{b}$ and $y_{L}^{*} = y_{L}/r_{b}$. Substituting Eq.~\eqref{eq:scales} into Eqs.~\eqref{eq:gov_dim}--\eqref{eq:source_dim} and dividing by $kT_{a}/r_{b}^{2}$ gives
\begin{align}
-\left( \frac{\partial^{2} T^{*}(x^{*},y^{*})}{\partial x^{*2}}
+ \frac{\partial^{2} T^{*}(x^{*},y^{*})}{\partial y^{*2}} \right)
&= q^{*}(x^{*},y^{*}), \qquad (x^{*},y^{*}) \in \Omega^{*}
\label{eq:gov_nondim} \\[2pt]
T^{*}(x^{*},y^{*}) &= 1, \qquad (x^{*},y^{*}) \in \partial \Omega^{*}
\label{eq:bc_nondim} \\[2pt]
q^{*}(x^{*},y^{*}) &= \Lambda \exp\!\left( -2\left[ (x^{*} - x_{L}^{*})^{2} + (y^{*} - y_{L}^{*})^{2} \right] \right)
\label{eq:source_nondim}
\end{align}
where $\Lambda = 2 Q_{\mathrm{total}}/(\pi k T_{a} h)$ is the dimensionless source amplitude. With the values in Table~\ref{tab:case_params}, $\Lambda \approx 12.99$ and $(x_{L}^{*}, y_{L}^{*}) = (-1.6, 0)$. Eqs.~\eqref{eq:gov_nondim}--\eqref{eq:source_nondim} thus define the thermal problem shared by all clients, on every geometry considered in this study (Section~\ref{sec:geometry_generation}). From here on, all quantities are nondimensional, and the asterisks are dropped for brevity.

\subsection{\emph{Fed-PIGANO: Federated Training of the Geometry-Aware Thermal Model with FedAvg-Adam}}
\label{sec:fedavg_adam}

\begin{figure}
    \centering
    \includegraphics[width=0.95\linewidth]{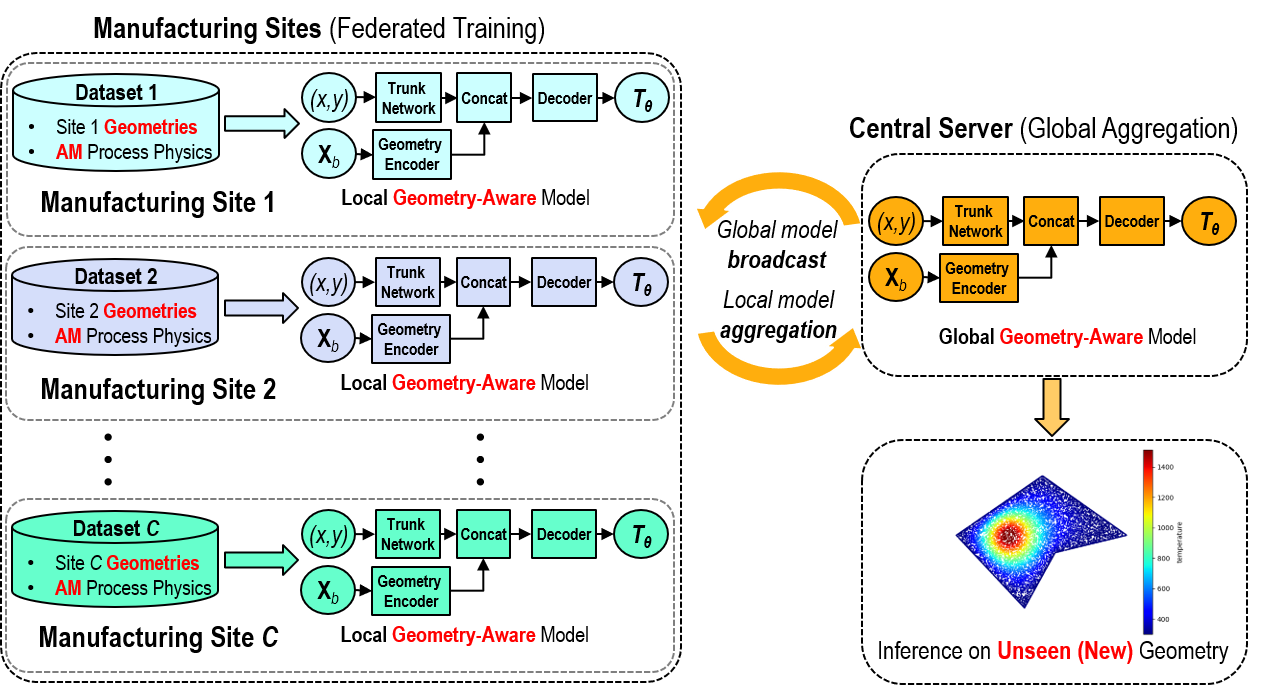}
    \caption{Fed-PIGANO framework. Each site $c \in \{1,\ldots,C\}$ trains a local model $\boldsymbol{\theta}_{c}$ that maps $\mathbf{X}_{b}$ and $(x,y)$ to $T_{\boldsymbol{\theta}}$, using only its own geometries. In each round, the server aggregates all model components into $\boldsymbol{\theta}^{r} = \sum_{c} w_{c}\,\boldsymbol{\theta}_{c}^{r}$ and broadcasts it back. The trained global model predicts thermal fields on unseen geometries.}
    \label{fig:Fed_GANO_Network}
\end{figure}

Fig.~\ref{fig:Fed_GANO_Network} illustrates the proposed Fed-PIGANO framework for the federated problem of Eq.~\eqref{eq:global_obj}. Each client trains a geometry-aware local model (Section~\ref{subsec:pigano_arch}) with a physics-informed loss (Section~\ref{sec:Fed_loss}) that uses only the governing equation and boundary condition of Section~\ref{sec:Fed_problem}. In each communication round, the server broadcasts the global parameters, each client trains on its own geometries, and the server aggregates and validates the updated parameters. All components of the local model, including the geometry encoder, are aggregated, so the global model learns from the geometries of all clients. Instead of the stochastic gradient descent of standard FedAvg~\citep{McMahan2017_Communication_FedAvg}, we follow \citet{Zhang__LuLu2025federated_SciML_functions_PDE_data_heterogeneity} and use Adam~\citep{Kingma2014_Adam_paper} for local updates, keeping each client's moment estimates $\mathbf{m}_{c}$ and $\mathbf{v}_{c}$ across rounds (FedAvg-Adam). Algorithm~\ref{alg:fedavg_adam} summarizes the procedure, and the three steps of a round are described below.

\textit{Broadcast and local training.} At the start of round $r$, each client sets $\boldsymbol{\theta}_{c} \leftarrow \boldsymbol{\theta}^{r-1}$, while $\mathbf{m}_{c}$ and $\mathbf{v}_{c}$ carry over from the previous round. The client then runs $E$ local epochs over $\mathcal{D}_{c}^{\mathrm{tr}}$. Let $t$ be the step counter of client $c$, which is not reset between rounds. At step $t$, the client takes a mini-batch $\mathcal{B} \subset \mathcal{D}_{c}^{\mathrm{tr}}$ of at most $B$ geometries, computes $\nabla_{c}^{t} = \nabla_{\boldsymbol{\theta}} \mathcal{L}(\boldsymbol{\theta}_{c};\mathcal{B})$ from Eq.~\eqref{eq:local_loss}, and updates
\begin{equation}
  \begin{aligned}
    \mathbf{m}_{c}^{t} &= \beta_{1}\,\mathbf{m}_{c}^{t-1} + (1-\beta_{1})\,\nabla_{c}^{t}, &
    \hat{\mathbf{m}}_{c}^{t} &= \mathbf{m}_{c}^{t}/(1-\beta_{1}^{t}), \\
    \mathbf{v}_{c}^{t} &= \beta_{2}\,\mathbf{v}_{c}^{t-1} + (1-\beta_{2})\,\nabla_{c}^{t} \odot \nabla_{c}^{t}, &
    \hat{\mathbf{v}}_{c}^{t} &= \mathbf{v}_{c}^{t}/(1-\beta_{2}^{t}), \\
    \boldsymbol{\theta}_{c} &\leftarrow \boldsymbol{\theta}_{c} - \eta\, \hat{\mathbf{m}}_{c}^{t} \oslash \left(\sqrt{\hat{\mathbf{v}}_{c}^{t}} + \varepsilon\right) & &
  \end{aligned}
  \label{eq:adam}
\end{equation}
where $\odot$, $\oslash$, and the square root act elementwise, $\eta$ is the learning rate, and $\beta_{1}$, $\beta_{2}$, $\varepsilon$ are the Adam constants. The moments are not reset, so the adaptive step size of each client reflects its full gradient history rather than only the current round. After $E$ epochs, the client holds the updated parameters $\boldsymbol{\theta}_{c}^{r}$.

\textit{Aggregation.} All clients participate in every round. The server aggregates their parameters as
\begin{equation}
  \boldsymbol{\theta}^{r} = \sum_{c=1}^{C} w_{c}\, \boldsymbol{\theta}_{c}^{r}
  \label{eq:fedavg_agg}
\end{equation}
The global model is thus a parameter average of local models trained on different sets of geometries. Only model parameters are communicated, and the moment estimates $\mathbf{m}_{c}$ and $\mathbf{v}_{c}$ stay at each client.

\begin{algorithm}[!t]
\small
\caption{Fed-PIGANO training with FedAvg-Adam}
\label{alg:fedavg_adam}
\begin{algorithmic}[1]
\Require Client sets $\{\mathcal{D}_{c}^{\mathrm{tr}}, \mathcal{D}_{c}^{\mathrm{val}}\}_{c=1}^{C}$; weights $w_{c}$; rounds $R$; local epochs $E$; batch size $B$; $\eta_{0}, \beta_{1}, \beta_{2}, \varepsilon$; loss weights $\lambda_{\mathrm{bc}}, \lambda_{\mathrm{pde}}$; scheduler settings (Table~\ref{tab:hyperparams})
\State Initialize $\boldsymbol{\theta}^{0} = (\boldsymbol{\theta}_{\mathrm{geo}}, \boldsymbol{\theta}_{\mathrm{tr}}, \boldsymbol{\theta}_{\mathrm{dec}})$ \Comment{Section~\ref{sec:impl_details}}
\State At every client: $\mathbf{m}_{c}, \mathbf{v}_{c} \leftarrow \mathbf{0}$, $t \leftarrow 0$; \; at the server: $\eta \leftarrow \eta_{0}$, $\mathcal{L}_{\mathrm{val}}^{*} \leftarrow +\infty$
\For{$r = 1, \ldots, R$}
  \For{each client $c = 1, \ldots, C$ \textbf{in parallel}}
    \State $\boldsymbol{\theta}_{c} \leftarrow \boldsymbol{\theta}^{r-1}$ 
    \Comment{Broadcast $\boldsymbol{\theta}^{r-1}$ and $\eta$; $\mathbf{m}_{c}, \mathbf{v}_{c}$ kept}
    \For{$e = 1, \ldots, E$}
      \For{each mini-batch $\mathcal{B} \subset \mathcal{D}_{c}^{\mathrm{tr}}$, $|\mathcal{B}| \le B$} \Comment{random partition}
        \State $t \leftarrow t + 1$
        \State $\nabla_{c}^{t} \leftarrow \nabla_{\boldsymbol{\theta}} \mathcal{L}(\boldsymbol{\theta}_{c};\mathcal{B})$ \Comment{Eq.~\eqref{eq:local_loss}}
        \State Update $\mathbf{m}_{c}, \mathbf{v}_{c}, \boldsymbol{\theta}_{c}$ \Comment{Eq.~\eqref{eq:adam}}
      \EndFor
    \EndFor
    \State Send $\boldsymbol{\theta}_{c}^{r} \leftarrow \boldsymbol{\theta}_{c}$ to the server
  \EndFor
  \State $\boldsymbol{\theta}^{r} \leftarrow \sum_{c} w_{c}\, \boldsymbol{\theta}_{c}^{r}$ \Comment{Aggregation, Eq.~\eqref{eq:fedavg_agg}}
  \State Each client returns $\mathcal{L}(\boldsymbol{\theta}^{r};\mathcal{D}_{c}^{\mathrm{val}})$ \Comment{Validation and model selection}
  \State $\mathcal{L}_{\mathrm{val}}^{r} \leftarrow \sum_{c} w_{c}\, \mathcal{L}(\boldsymbol{\theta}^{r};\mathcal{D}_{c}^{\mathrm{val}})$ \Comment{Eq.~\eqref{eq:val_loss}}
  \If{$\mathcal{L}_{\mathrm{val}}^{r} < \mathcal{L}_{\mathrm{val}}^{*}$}
    \State $\boldsymbol{\theta}^{*} \leftarrow \boldsymbol{\theta}^{r}$; \; $r^{*} \leftarrow r$; \; $\mathcal{L}_{\mathrm{val}}^{*} \leftarrow \mathcal{L}_{\mathrm{val}}^{r}$ \Comment{Eq.~\eqref{eq:checkpoint}}
  \EndIf
  \State $\eta \leftarrow \textsc{ReduceLROnPlateau}(\eta, \mathcal{L}_{\mathrm{val}}^{r})$ \Comment{Shared by all clients}
\EndFor
\State \Return $\boldsymbol{\theta}^{*}$
\end{algorithmic}
\end{algorithm}

\textit{Validation and model selection.} After aggregation, each client evaluates $\boldsymbol{\theta}^{r}$ on its validation set and returns the scalar loss to the server, which forms
\begin{equation}
  \mathcal{L}_{\mathrm{val}}^{r} = \sum_{c=1}^{C} w_{c}\, \mathcal{L}\!\left(\boldsymbol{\theta}^{r};\mathcal{D}_{c}^{\mathrm{val}}\right)
  \label{eq:val_loss}
\end{equation}
The server uses $\mathcal{L}_{\mathrm{val}}^{r}$ for two purposes. First, it controls a single learning rate $\eta$ shared by all clients. The rate starts at $\eta_{0}$ and is reduced by a fixed factor, down to a floor $\eta_{\min}$, with \texttt{ReduceLROnPlateau} when $\mathcal{L}_{\mathrm{val}}^{r}$ has not decreased by a relative threshold for a set number of rounds (the patience). After each reduction, the rate is held fixed for a cooldown period before it can be reduced again. Second, it selects the global checkpoint with the lowest validation loss,
\begin{equation}
  r^{*} = \arg\min_{r \in \{1,\ldots,R\}} \mathcal{L}_{\mathrm{val}}^{r}, \qquad \boldsymbol{\theta}^{*} = \boldsymbol{\theta}^{r^{*}}
  \label{eq:checkpoint}
\end{equation}
Model selection thus requires no reference solutions, as $\mathcal{L}$ uses only the governing equation and boundary condition (Section~\ref{sec:Fed_loss}). The values of all training settings are given in Section~\ref{sec:impl_details}.

\subsection{\emph{Local Geometry-Aware Model (PI-GANO)}}
\label{subsec:pigano_arch}
We adopt the geometry-only variant of PI-GANO~\citep{Zhong_Meidani2025_PI-GANO_both_Geometry_parameters} as the local model of Fed-PIGANO at each client, since it predicts PDE solutions on varying domains given the geometry. Hereafter, we refer to this variant as PI-GANO. The model has three trainable components, shown for each client in Fig.~\ref{fig:Fed_GANO_Network}, a geometry encoder with parameters $\boldsymbol{\theta}_{\mathrm{geo}}$, a trunk network with parameters $\boldsymbol{\theta}_{\mathrm{tr}}$, and a decoder with parameters $\boldsymbol{\theta}_{\mathrm{dec}}$. We write $\boldsymbol{\theta} = (\boldsymbol{\theta}_{\mathrm{geo}}, \boldsymbol{\theta}_{\mathrm{tr}}, \boldsymbol{\theta}_{\mathrm{dec}})$ for the full set of weights and biases.

Following \citet{Zhong_Meidani2025_PI-GANO_both_Geometry_parameters}, the geometry encoder $\mathcal{F}_{\mathrm{geo}}$ is a four-layer fully connected network with widths $2 \to d_{h} \to d_{h} \to d_{h} \to d_{h}$, $\tanh$ activations on the hidden layers, and a linear output layer. It is applied with shared weights to every boundary point, and the resulting features are averaged into the geometry embedding
\begin{equation}
  \mathbf{z}_{\Omega} = \frac{1}{n_{b}} \sum_{(x,y) \in \mathbf{X}_{b}} \mathcal{F}_{\mathrm{geo}}\!\left([x,\,y]^{\top}\right) \in \mathbb{R}^{d_{h}}
  \label{eq:dg}
\end{equation}
Mean pooling makes $\mathbf{z}_{\Omega}$ independent of the ordering of the boundary points \citep{Qi_Su__Guibas2017_PointNet, Saimon2026_GeoPIPN}, and the same embedding is used for every query point on that geometry. The trunk network is a single linear layer that lifts the query point to $\mathbf{u}(x,y) = \mathbf{W}^{\mathrm{tr}}[x,\,y]^{\top} + \mathbf{b}^{\mathrm{tr}} \in \mathbb{R}^{d_{h}}$. The two features are then concatenated and passed to the decoder $\mathcal{F}_{\mathrm{dec}}$, a fully connected network of the same form with widths $2d_{h} \to 2d_{h} \to 2d_{h} \to 2d_{h} \to 1$:
\begin{equation}
  T_{\boldsymbol{\theta}}(x,y;\mathbf{X}_{b}) = \mathcal{F}_{\mathrm{dec}}\!\left(\left[\mathbf{u}(x,y) \,\|\, \mathbf{z}_{\Omega}\right]\right)
  \label{eq:trunk}
\end{equation}
where $\|$ denotes concatenation. The prediction therefore depends on both the query location and the shape of the domain. All three components are trained locally with the physics-informed loss of Section~\ref{sec:Fed_loss} and aggregated at the server by Eq.~\eqref{eq:fedavg_agg}.

\subsection{\emph{Physics-Informed Local Loss Formulation}}
\label{sec:Fed_loss}
We train PI-GANO with a physics-informed loss derived only from the governing equation and boundary condition of Section~\ref{sec:Fed_problem}, without temperature labels from either simulations or sensors. FEM solutions serve only as reference fields for evaluation (Section~\ref{sec:eval_metrics}). For a single geometry $g$, the loss combines a boundary-condition term and a PDE-residual term,
\begin{equation}
  \ell(\boldsymbol{\theta}; g) = \lambda_{\mathrm{bc}}\,\ell_{\mathrm{bc}}(\boldsymbol{\theta}; g) + \lambda_{\mathrm{pde}}\,\ell_{\mathrm{pde}}(\boldsymbol{\theta}; g)
  \label{eq:geom_loss}
\end{equation}
where $\lambda_{\mathrm{bc}}$ and $\lambda_{\mathrm{pde}}$ are loss weights, both set to 1 (Table~\ref{tab:hyperparams}). The boundary term enforces the Dirichlet condition~\eqref{eq:bc_nondim} on the boundary point set:
\begin{equation}
  \ell_{\mathrm{bc}}(\boldsymbol{\theta}; g) = \frac{1}{n_{b}} \sum_{(x,y) \in \mathbf{X}_{b}^{g}}
  \left[ T_{\boldsymbol{\theta}}(x,y;\mathbf{X}_{b}^{g}) - 1 \right]^{2}
  \label{eq:loss_bc}
\end{equation}
The PDE term enforces Eq.~\eqref{eq:gov_nondim} at every point of the full point set $\mathbf{X}^{g}$, including the boundary points:
\begin{equation}
  \ell_{\mathrm{pde}}(\boldsymbol{\theta}; g) = \frac{1}{N_{\mathrm{pts}}} \sum_{(x,y) \in \mathbf{X}^{g}}
  \left[ \mathcal{R}_{\boldsymbol{\theta}}(x,y;\mathbf{X}_{b}^{g}) \right]^{2}
  \label{eq:loss_pde}
\end{equation}
where the pointwise residual is
\begin{equation}
  \mathcal{R}_{\boldsymbol{\theta}}(x,y;\mathbf{X}_{b}) =
  -\left( \frac{\partial^{2} T_{\boldsymbol{\theta}}}{\partial x^{2}}
  + \frac{\partial^{2} T_{\boldsymbol{\theta}}}{\partial y^{2}} \right)
  - \Lambda \exp\!\left( -2\left[ (x - x_{L})^{2} + (y - y_{L})^{2} \right] \right)
  \label{eq:residual}
\end{equation}
with $\Lambda$ and $(x_{L}, y_{L})$ as given in Section~\ref{subsec:nondim}. The second derivatives are obtained by automatic differentiation of $T_{\boldsymbol{\theta}}$ with respect to $x$ and $y$.

For any set $\mathcal{S}$ of geometries, such as a mini-batch, a training set, or a validation set, we define the loss as the average over its members,
\begin{equation}
  \mathcal{L}(\boldsymbol{\theta};\mathcal{S}) = \frac{1}{|\mathcal{S}|} \sum_{g \in \mathcal{S}} \ell(\boldsymbol{\theta}; g)
  \label{eq:local_loss}
\end{equation}
The residual~\eqref{eq:residual} has the same operator and source term at every client, and client data enter the loss only through the geometry. All local models therefore learn the same process physics on different sets of geometries, which makes aggregating their parameters by Eq.~\eqref{eq:fedavg_agg} meaningful.

\section{Experimental Setup}
\label{sec:Fed_experimental_setup}
We evaluate Fed-PIGANO on the 2D AM thermal problem of Section~\ref{sec:Fed_problem}. This section describes the geometry generation (Section~\ref{sec:geometry_generation}), the client scenarios (Section~\ref{sec:client_distributions}), the Local-only baseline and the Centralized reference (Section~\ref{sec:local_baseline}), the implementation and the seven replications (Section~\ref{sec:impl_details}), and the evaluation metrics (Section~\ref{sec:eval_metrics}).

\subsection{\emph{Random Geometry Generation and Discretization}}
\label{sec:geometry_generation}
\begin{figure}[htbp]
  \centering
  \includegraphics[width=0.95\linewidth]{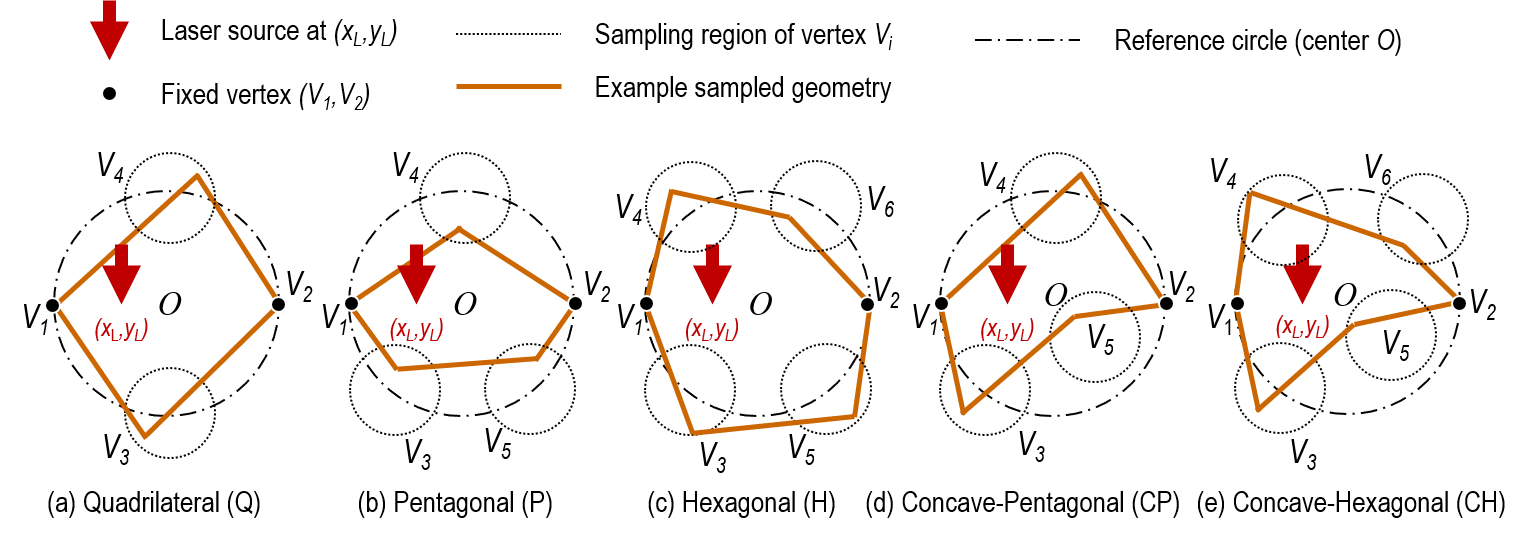}
  \caption{Geometry generation scheme for the five families. Orange polygons show one example geometry per family.}
  \label{fig:geometry_generation_scheme}
\end{figure}

Each domain $\Omega$ belongs to one of $F=5$ polygonal geometry families. The Quadrilateral (Q), Pentagonal (P), and Hexagonal (H) families are convex, and the Concave-Pentagonal (CP) and Concave-Hexagonal (CH) families are concave. Fig.~\ref{fig:geometry_generation_scheme} shows how we generate the geometries of each family. The vertices $V_{1}$ and $V_{2}$ are fixed where the domain centerline meets a reference circle of radius 10~mm centered at $O$, so that every geometry has the same extent along the centerline and contains the fixed source position (Section~\ref{subsec:thermal_model}). Each remaining vertex $V_{i}$ is sampled uniformly within a circle of radius 3~mm around its nominal position. In the concave families, the sampling region of $V_{5}$ lies inside the reference circle, which creates a reentrant corner. Each family comprises 300 distinct geometries, which are randomly partitioned among the $C=3$ clients, so that each client holds a local pool of 100 geometries per family (500 in total).

For each geometry, we solve the thermal problem of Eqs.~\eqref{eq:gov_dim}--\eqref{eq:source_dim} with the finite element method (FEM) using the MATLAB PDE solver, with a maximum element size of $r_{b}/28$. The point set $\mathbf{X}$ of each geometry (Section~\ref{sec:Fed_problem_statement}) is a random sample of $n_{b} = 500$ boundary nodes and $n_{\mathrm{int}} = 1{,}000$ interior nodes of the mesh, so that $N_{\mathrm{pts}} = 1{,}500$. The mesh thus supplies the training point locations, while the FEM temperatures at these nodes serve only as reference values for evaluation (Section~\ref{sec:eval_metrics}).

\subsection{\emph{Client Geometry Distributions and Scenarios}}
\label{sec:client_distributions}

To emulate manufacturing sites whose design geometries cover different parts of the geometry space (Section~\ref{sec:Fed_problem_statement}), we construct four scenarios in which the clients' training geometries differ in family composition. From its local pool of 500 geometries (Section~\ref{sec:geometry_generation}), each client draws disjoint training, validation, and test sets in each replication of the experiment (Section~\ref{sec:impl_details}). The test set $\mathcal{D}_{c}^{\mathrm{te}}$ contains 10 geometries per family ($G = 50$). Within a replication, it is drawn once per client and reused in every scenario, so differences in test performance across scenarios reflect only the training and validation sets. The training set $\mathcal{D}_{c}^{\mathrm{tr}}$ always contains $n_{c} = 50$ geometries, and the validation set $\mathcal{D}_{c}^{\mathrm{val}}$ contains 25 geometries with the same family proportions. The scenarios differ only in how the 50 training geometries are split across the five families. Table~\ref{tab:Fed_scenarios} lists the family counts of all sets. 

\begin{table}[htbp]
    \footnotesize
    \centering
    \caption{Per-client family counts (Q, P, H, CP, CH) of the training, validation, and test sets in each scenario. Larger $\delta$ (Eq.~\eqref{eq:delta}) indicates stronger heterogeneity across clients.}
    \label{tab:Fed_scenarios}
    \renewcommand{\arraystretch}{0.80}
    \begin{tabular}{llccc}
        \toprule
        Scenario & Split & Client 1 & Client 2 & Client 3 \\
        \midrule
        \multirow{3}{*}{S1 ($\delta = 0.000$)}
        & Train (50) & (10,\,10,\,10,\,10,\,10) & (10,\,10,\,10,\,10,\,10) & (10,\,10,\,10,\,10,\,10) \\
        & Val (25)   & (\phantom{0}5,\,\phantom{0}5,\,\phantom{0}5,\,\phantom{0}5,\,\phantom{0}5) & (\phantom{0}5,\,\phantom{0}5,\,\phantom{0}5,\,\phantom{0}5,\,\phantom{0}5) & (\phantom{0}5,\,\phantom{0}5,\,\phantom{0}5,\,\phantom{0}5,\,\phantom{0}5) \\
        & Test (50)  & (10,\,10,\,10,\,10,\,10) & (10,\,10,\,10,\,10,\,10) & (10,\,10,\,10,\,10,\,10) \\
        \midrule
        \multirow{3}{*}{S2 ($\delta = 0.198$)}
        & Train (50) & (22,\,16,\,\phantom{0}6,\,\phantom{0}4,\,\phantom{0}2) & (\phantom{0}4,\,10,\,18,\,12,\,\phantom{0}6) & (\phantom{0}4,\,\phantom{0}4,\,\phantom{0}6,\,14,\,22) \\
        & Val (25)   & (11,\,\phantom{0}8,\,\phantom{0}3,\,\phantom{0}2,\,\phantom{0}1) & (\phantom{0}2,\,\phantom{0}5,\,\phantom{0}9,\,\phantom{0}6,\,\phantom{0}3) & (\phantom{0}2,\,\phantom{0}2,\,\phantom{0}3,\,\phantom{0}7,\,11) \\
        & Test (50)  & (10,\,10,\,10,\,10,\,10) & (10,\,10,\,10,\,10,\,10) & (10,\,10,\,10,\,10,\,10) \\
        \midrule
        \multirow{3}{*}{S3 ($\delta = 0.449$)}
        & Train (50) & (26,\,20,\,\phantom{0}4,\,\phantom{0}0,\,\phantom{0}0) & (\phantom{0}4,\,\phantom{0}8,\,22,\,12,\,\phantom{0}4) & (\phantom{0}0,\,\phantom{0}2,\,\phantom{0}4,\,18,\,26) \\
        & Val (25)   & (13,\,10,\,\phantom{0}2,\,\phantom{0}0,\,\phantom{0}0) & (\phantom{0}2,\,\phantom{0}4,\,11,\,\phantom{0}6,\,\phantom{0}2) & (\phantom{0}0,\,\phantom{0}1,\,\phantom{0}2,\,\phantom{0}9,\,13) \\
        & Test (50)  & (10,\,10,\,10,\,10,\,10) & (10,\,10,\,10,\,10,\,10) & (10,\,10,\,10,\,10,\,10) \\
        \midrule
        \multirow{3}{*}{S4 ($\delta = 0.768$)}
        & Train (50) & (30,\,20,\,\phantom{0}0,\,\phantom{0}0,\,\phantom{0}0) & (\phantom{0}0,\,10,\,30,\,10,\,\phantom{0}0) & (\phantom{0}0,\,\phantom{0}0,\,\phantom{0}0,\,20,\,30) \\
        & Val (25)   & (15,\,10,\,\phantom{0}0,\,\phantom{0}0,\,\phantom{0}0) & (\phantom{0}0,\,\phantom{0}5,\,15,\,\phantom{0}5,\,\phantom{0}0) & (\phantom{0}0,\,\phantom{0}0,\,\phantom{0}0,\,10,\,15) \\
        & Test (50)  & (10,\,10,\,10,\,10,\,10) & (10,\,10,\,10,\,10,\,10) & (10,\,10,\,10,\,10,\,10) \\
        \bottomrule
    \end{tabular}%
\end{table}

We quantify the difference in family composition with the Jensen--Shannon divergence generalized to $C$ distributions \citep{Lin1991_divergence_Shannon_entropy}. Let $n_{c}^{f}$ be the number of training geometries of family $f$ at client $c$, and let $\mathbf{p}_{c} = (p_{c}^{1},\ldots,p_{c}^{F})$, with $p_{c}^{f} = n_{c}^{f}/n_{c}$, be the family-proportion vector of client $c$. Let $\bar{\mathbf{p}} = C^{-1}\sum_{c=1}^{C}\mathbf{p}_{c}$ be the global mixture. With equal client weights, this divergence is bounded above by $\ln C$ \citep{Lin1991_divergence_Shannon_entropy}, and we normalize it by this bound as
\begin{equation}
  \delta = \frac{1}{\ln C}\left[ H(\bar{\mathbf{p}}) - \frac{1}{C}\sum_{c=1}^{C} H(\mathbf{p}_{c}) \right] \in [0,\,1]
  \label{eq:delta}
\end{equation}
where $H(\mathbf{p}) = -\sum_{f=1}^{F} p^{f}\ln p^{f}$ is the Shannon entropy, with $0\ln 0 = 0$. The value of $\delta$ is 0 when all clients have the same family proportions and 1 when no two clients share a family. Here, $\delta$ measures heterogeneity in family composition, not the similarity between individual shapes. In S1, every client has 10 training geometries per family. From S2 to S4, Client~1 trains increasingly on the convex families Q and P, Client~2 on H, and Client~3 on the concave families CP and CH. In S4, the only shared families are P (Clients~1 and~2) and CP (Clients~2 and~3). In every scenario, each family has 30 training geometries in total, so the global mixture $\bar{\mathbf{p}}$ is uniform and changes in $\delta$ reflect only differences between clients. The divergence increases from $\delta = 0$ (S1) to $0.198$ (S2), $0.449$ (S3), and $0.768$ (S4).

\subsection{\emph{Local-Only Baseline and Centralized Reference}}
\label{sec:local_baseline}
The Local-only baseline represents the setting of Fig.~\ref{fig:motivation_federated_geometry_generalization}b, in which each site trains a geometry-aware model on its own geometries only. Each client $c$ trains its own PI-GANO model $\boldsymbol{\theta}_{c}^{\mathrm{loc}}$ on $\mathcal{D}_{c}^{\mathrm{tr}}$ without communication, using the same architecture (Section~\ref{subsec:pigano_arch}), loss (Eq.~\eqref{eq:local_loss}), initialization, optimizer, mini-batch size, and learning-rate schedule as Fed-PIGANO. Training runs for $E_{\max}$ epochs. After each epoch $e$, the model $\boldsymbol{\theta}_{c}^{\mathrm{loc},e}$ is evaluated on $\mathcal{D}_{c}^{\mathrm{val}}$, and this validation loss both drives the learning-rate schedule and selects the checkpoint,
\begin{equation}
  e_{c}^{*} = \arg\min_{e \in \{1,\ldots,E_{\max}\}} \mathcal{L}\!\left(\boldsymbol{\theta}_{c}^{\mathrm{loc},e};\mathcal{D}_{c}^{\mathrm{val}}\right),
  \qquad
  \boldsymbol{\theta}_{c}^{\mathrm{loc}} = \boldsymbol{\theta}_{c}^{\mathrm{loc},e_{c}^{*}}
  \label{eq:local_checkpoint}
\end{equation}
which mirrors the checkpoint selection of Fed-PIGANO (Eq.~\eqref{eq:checkpoint}) at a single client. We set $E_{\max} = R \times E$, so each Local-only model makes the same number of passes over its training set as each client does in Fed-PIGANO. The scheduler settings in Table~\ref{tab:hyperparams} are equivalent in the same sense, since a patience of 100 rounds and a cooldown of 20 rounds correspond to 500 and 100 epochs at $E = 5$. We also train a Centralized reference, which represents the pooling setting of Fig.~\ref{fig:motivation_federated_geometry_generalization}c. It is a single PI-GANO model $\boldsymbol{\theta}^{\mathrm{cen}}$ trained like the Local-only models, using the pooled training and validation sets of all clients. Centralized requires all geometries in one place, which our setting rules out, so it serves only as a reference, not as a competing method.

\subsection{\emph{Implementation Details}}
\label{sec:impl_details}
Table~\ref{tab:hyperparams} lists the training settings of the three models. Local-only and Fed-PIGANO differ only in federation (Section~\ref{sec:local_baseline}), so the performance differences between them reported in Section~\ref{sec:Fed_results_discussion} can be attributed to federated training. The PI-GANO architecture follows \citet{Zhong_Meidani2025_PI-GANO_both_Geometry_parameters} (Section~\ref{subsec:pigano_arch}). We repeat the experiment seven times with random seeds fixed in advance and report all seven replications. The seed of a replication sets the split of each client's geometries into training, validation, and test sets with the counts of Table~\ref{tab:Fed_scenarios}, the initial weights, and the mini-batch order. Within a replication, every Local-only model, the initial global model $\boldsymbol{\theta}^{0}$ of Fed-PIGANO, and the Centralized model start from identical parameters. We implemented all experiments in PyTorch 2.8.0 and ran them on a single NVIDIA GeForce RTX 4090 GPU. One training run takes about 13~min for a Local-only model, 36~min for Centralized, and 39~min for Fed-PIGANO, whose three clients are simulated one after another.
\begin{table}[htbp]
    \centering
    \footnotesize
    \caption{Training settings of the three models. Shared settings span all three columns.}
    \label{tab:hyperparams}
    \renewcommand{\arraystretch}{0.90}
    \begin{tabular*}{\textwidth}{@{\extracolsep{\fill}}lccc@{}}
        \toprule
        Setting & Local-only & Fed-PIGANO & Centralized \\
        \midrule
        Training / validation geometries & $50$ / $25$ & $50$ / $25$ per client & $150$ / $75$ (pooled) \\
        Hidden width $d_{h}$ & \multicolumn{3}{c}{$128$} \\
        Weight initialization & \multicolumn{3}{c}{Glorot uniform ($\tanh$ gain), zero biases} \\
        Optimizer & \multicolumn{3}{c}{Adam ($\eta_{0}=10^{-3}$, $\beta_{1}=0.9$, $\beta_{2}=0.999$, $\varepsilon=10^{-8}$)} \\
        Learning-rate schedule & \multicolumn{3}{c}{\texttt{ReduceLROnPlateau} (factor $0.5$, relative threshold $5\times10^{-3}$)} \\
        Patience / cooldown & $500$ / $100$ epochs & $100$ / $20$ rounds & $500$ / $100$ epochs \\
        Learning-rate floor $\eta_{\min}$ & \multicolumn{3}{c}{$10^{-6}$} \\
        Mini-batch size $B$ (geometries) & \multicolumn{3}{c}{$8$} \\
        Loss weights & \multicolumn{3}{c}{$\lambda_{\mathrm{bc}} = \lambda_{\mathrm{pde}} = 1$} \\
        Training length & $E_{\max} = 12{,}000$ epochs & $R = 2{,}400$ rounds, $E = 5$ & $E_{\max} = 12{,}000$ epochs \\
        Model selection & Eq.~\eqref{eq:local_checkpoint} & Eq.~\eqref{eq:checkpoint} & Eq.~\eqref{eq:local_checkpoint} \\
        \bottomrule
    \end{tabular*}
\end{table}

\subsection{\emph{Evaluation Metrics}}
\label{sec:eval_metrics}
We evaluate all three models with two metrics, one for accuracy against the FEM reference and one for consistency with the governing equation. The global Fed-PIGANO model $\boldsymbol{\theta}^{*}$ and the Centralized model $\boldsymbol{\theta}^{\mathrm{cen}}$ are tested on the test set $\mathcal{D}_{c}^{\mathrm{te}}$ of every client ($G = 50$ geometries, Section~\ref{sec:client_distributions}), and each Local-only model $\boldsymbol{\theta}_{c}^{\mathrm{loc}}$ on its own client's test set. For a test geometry $g$, let $T^{\mathrm{FEM}}_{g}(x,y)$ be the FEM reference temperature at a point $(x,y) \in \mathbf{X}^{g}$. The relative $\ell_2$ error on $g$, which is the same in dimensional and nondimensional units, is
\begin{equation}
  \epsilon_{\mathrm{rel}} =
  \sqrt{\frac{\sum_{(x,y) \in \mathbf{X}^{g}} \left[ T_{\boldsymbol{\theta}}(x,y;\mathbf{X}_{b}^{g}) - T^{\mathrm{FEM}}_{g}(x,y) \right]^{2}}
             {\sum_{(x,y) \in \mathbf{X}^{g}} \left[ T^{\mathrm{FEM}}_{g}(x,y) \right]^{2}}}
  \label{eq:rel_l2}
\end{equation}
The PDE-residual RMSE on $g$ is
\begin{equation}
  \epsilon_{\mathrm{PDE}} =
  \sqrt{\frac{1}{N_{\mathrm{pts}}} \sum_{(x,y) \in \mathbf{X}^{g}} \left[ \mathcal{R}_{\boldsymbol{\theta}}(x,y;\mathbf{X}_{b}^{g}) \right]^{2}}
  \label{eq:pde_rmse}
\end{equation}
where $\mathcal{R}_{\boldsymbol{\theta}}$ is the nondimensional residual of Eq.~\eqref{eq:residual} and $\boldsymbol{\theta}$ is the evaluated model. Lower values are better for both metrics.

\section{Results and Discussion}
\label{sec:Fed_results_discussion}

We compare Fed-PIGANO with the Local-only baseline at three levels of detail, with Centralized as a pooled-training reference. Section~\ref{sec:results_mean} reports the mean performance over seven replications of the experiment and how it changes with the heterogeneity $\delta$. Section~\ref{sec:results_geometry} examines one of these replications on individual test geometries, and Section~\ref{sec:results_fields} shows predicted thermal fields from the same replication.

\subsection{\emph{Mean Prediction Accuracy on Unseen Test Geometries over Seven Replications}}
\label{sec:results_mean}
Table~\ref{tab:results} reports the two metrics of Section~\ref{sec:eval_metrics} for every scenario and client. The experiment is replicated seven times (Section~\ref{sec:impl_details}), and each replication draws a new random split of every client's geometries into training, validation, and test sets. Within a replication, each metric is averaged over the $G = 50$ test geometries of a client. The table gives the mean and standard deviation of these averages over the seven replications. In S1, where all clients have identical geometry-family proportions, Local-only exhibits approximately 2.3 times the relative error and 1.9--2.0 times the PDE-residual RMSE of Fed-PIGANO. This improvement is consistent with the additional geometric diversity available through federation, as Fed-PIGANO learns from 150 training geometries compared with 50 at each individual site. The advantage persists across S2--S4, with Fed-PIGANO achieving lower mean values for both metrics in all 12 scenario--client cases. Centralized pools the geometries of all clients, which the setting of this study does not allow (Section~\ref{sec:local_baseline}), so Table~\ref{tab:results} includes it only as a reference for pooled training.
\begin{table}[!t]
    \centering
    \footnotesize
    \caption{Relative $\ell_2$ error and PDE-residual RMSE on unseen test geometries for each scenario and client, as mean $\pm$ standard deviation over seven replications. Bold marks the lower mean of Local-only and Fed-PIGANO.}
    \label{tab:results}
    \setlength{\tabcolsep}{3.5pt}
    \renewcommand{\arraystretch}{0.85}
    \begin{tabular*}{\textwidth}{@{\extracolsep{\fill}}lcccccc@{}}
        \toprule
        & \multicolumn{2}{c}{Client 1} & \multicolumn{2}{c}{Client 2} & \multicolumn{2}{c}{Client 3} \\
        \cmidrule(lr){2-3}\cmidrule(lr){4-5}\cmidrule(lr){6-7}
        Method
        & \makecell{$\epsilon_{\mathrm{rel}}\downarrow$\\(\%)} & \makecell{$\epsilon_{\mathrm{PDE}}\downarrow$\\($\times10^{-2}$)}
        & \makecell{$\epsilon_{\mathrm{rel}}\downarrow$\\(\%)} & \makecell{$\epsilon_{\mathrm{PDE}}\downarrow$\\($\times10^{-2}$)}
        & \makecell{$\epsilon_{\mathrm{rel}}\downarrow$\\(\%)} & \makecell{$\epsilon_{\mathrm{PDE}}\downarrow$\\($\times10^{-2}$)} \\
        \midrule
        \multicolumn{7}{@{}l}{S1 ($\delta = 0.000$)} \\
        \quad Local-only & $4.36 \pm 0.41$ & $6.55 \pm 0.77$ & $4.25 \pm 0.53$ & $6.88 \pm 0.65$ & $4.34 \pm 0.40$ & $6.43 \pm 0.65$ \\
        \quad Fed-PIGANO & $\mathbf{1.91} \pm 0.17$ & $\mathbf{3.34} \pm 0.25$ & $\mathbf{1.83} \pm 0.15$ & $\mathbf{3.40} \pm 0.35$ & $\mathbf{1.89} \pm 0.22$ & $\mathbf{3.45} \pm 0.28$ \\
        \quad \textit{Centralized}$^{\dagger}$ & $\mathit{1.91 \pm 0.10}$ & $\mathit{2.77 \pm 0.15}$ & $\mathit{1.82 \pm 0.10}$ & $\mathit{2.79 \pm 0.19}$ & $\mathit{1.89 \pm 0.15}$ & $\mathit{2.87 \pm 0.21}$ \\
        \midrule
        \multicolumn{7}{@{}l}{S2 ($\delta = 0.198$)} \\
        \quad Local-only & $4.73 \pm 0.33$ & $7.86 \pm 0.71$ & $4.30 \pm 0.45$ & $7.01 \pm 0.54$ & $4.80 \pm 0.51$ & $7.33 \pm 0.46$ \\
        \quad Fed-PIGANO & $\mathbf{1.83} \pm 0.08$ & $\mathbf{3.37} \pm 0.20$ & $\mathbf{1.90} \pm 0.13$ & $\mathbf{3.35} \pm 0.17$ & $\mathbf{1.92} \pm 0.15$ & $\mathbf{3.45} \pm 0.38$ \\
        \quad \textit{Centralized}$^{\dagger}$ & $\mathit{1.80 \pm 0.11}$ & $\mathit{2.81 \pm 0.21}$ & $\mathit{1.85 \pm 0.05}$ & $\mathit{2.76 \pm 0.10}$ & $\mathit{1.87 \pm 0.12}$ & $\mathit{2.87 \pm 0.23}$ \\
        \midrule
        \multicolumn{7}{@{}l}{S3 ($\delta = 0.449$)} \\
        \quad Local-only & $7.55 \pm 0.81$ & $13.64 \pm 4.52$ & $4.41 \pm 0.48$ & $7.09 \pm 0.94$ & $6.42 \pm 0.82$ & $9.20 \pm 0.97$ \\
        \quad Fed-PIGANO & $\mathbf{1.90} \pm 0.15$ & $\mathbf{3.47} \pm 0.27$ & $\mathbf{1.98} \pm 0.21$ & $\mathbf{3.59} \pm 0.20$ & $\mathbf{2.03} \pm 0.12$ & $\mathbf{3.63} \pm 0.37$ \\
        \quad \textit{Centralized}$^{\dagger}$ & $\mathit{1.86 \pm 0.12}$ & $\mathit{2.77 \pm 0.21}$ & $\mathit{1.91 \pm 0.14}$ & $\mathit{2.78 \pm 0.19}$ & $\mathit{1.90 \pm 0.11}$ & $\mathit{2.85 \pm 0.23}$ \\
        \midrule
        \multicolumn{7}{@{}l}{S4 ($\delta = 0.768$)} \\
        \quad Local-only & $10.16 \pm 0.82$ & $16.94 \pm 3.41$ & $4.96 \pm 0.24$ & $8.08 \pm 0.86$ & $20.46 \pm 2.04$ & $48.76 \pm 6.69$ \\
        \quad Fed-PIGANO & $\mathbf{1.99} \pm 0.16$ & $\mathbf{3.59} \pm 0.39$ & $\mathbf{2.00} \pm 0.22$ & $\mathbf{3.58} \pm 0.28$ & $\mathbf{2.04} \pm 0.20$ & $\mathbf{3.63} \pm 0.26$ \\
        \quad \textit{Centralized}$^{\dagger}$ & $\mathit{1.88 \pm 0.05}$ & $\mathit{2.82 \pm 0.15}$ & $\mathit{1.89 \pm 0.14}$ & $\mathit{2.77 \pm 0.18}$ & $\mathit{1.96 \pm 0.14}$ & $\mathit{2.92 \pm 0.31}$ \\
        \bottomrule
    \end{tabular*}
    \par\vspace{3pt}
    \parbox{\textwidth}{\footnotesize $\downarrow$\,Lower is better. $^{\dagger}$Reference trained on the pooled geometries of all clients, shown in italics and not part of the bold comparison.}
\end{table}

\begin{figure}[htbp]
    \centering
    \includegraphics[width=0.55\linewidth]{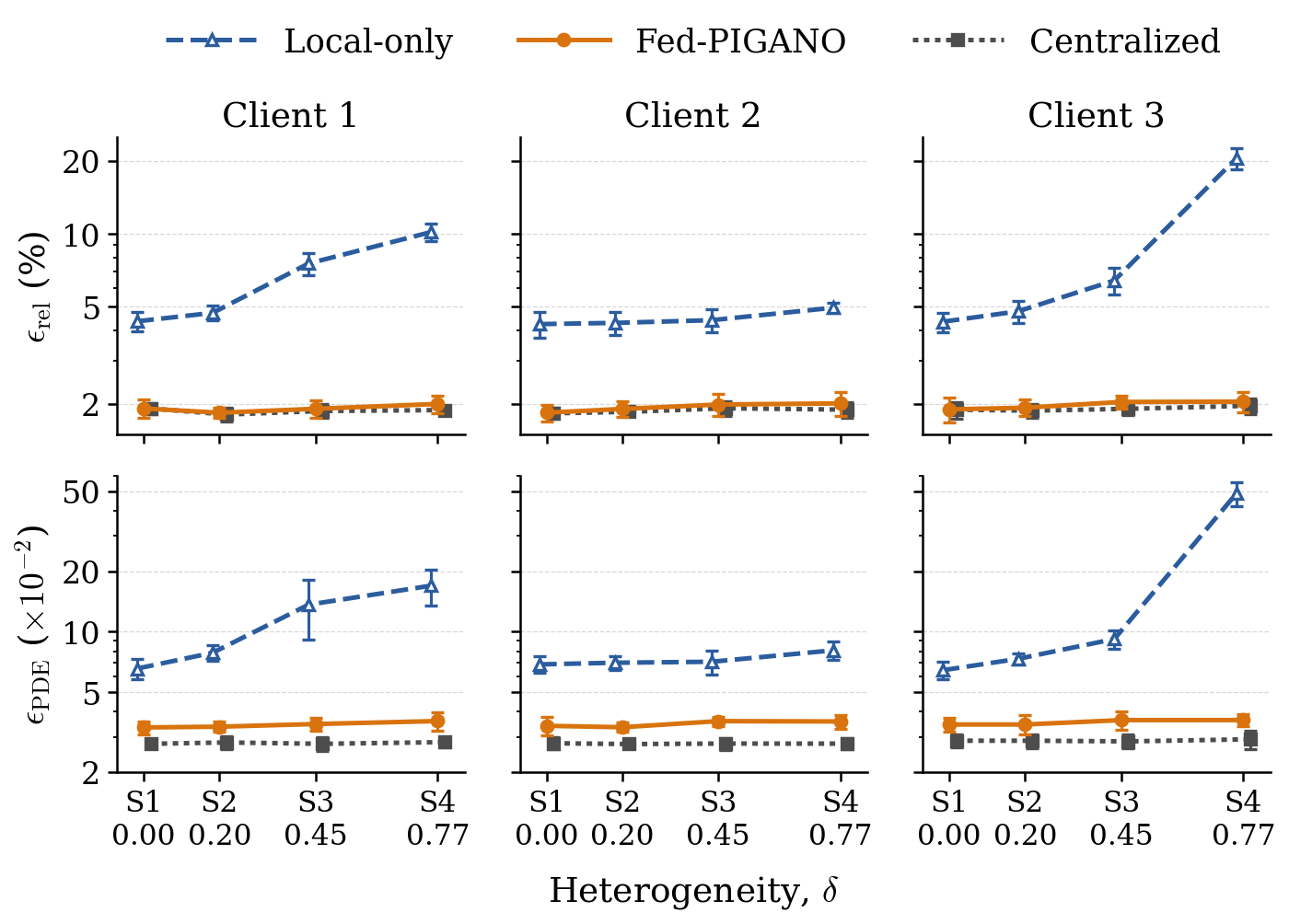}
    \caption{Mean test-set $\epsilon_{\mathrm{rel}}$ (top) and $\epsilon_{\mathrm{PDE}}$ (bottom) versus $\delta$ for each client (columns). Error bars show $\pm 1$ standard deviation over the seven replications.}
    \label{fig:error_vs_delta}
\end{figure}

We next examine how the performance on unseen geometries changes as the family compositions of the clients diverge. Fig.~\ref{fig:error_vs_delta} plots the means of Table~\ref{tab:results} against $\delta$. Within a replication, each client keeps the same test set in all four scenarios, so any change along $\delta$ comes only from the training and validation sets. In every scenario and for both metrics, Fed-PIGANO stays below Local-only, and their error bars do not overlap. Both Local-only metrics grow with $\delta$ for every client. The growth is largest for Client~3, which trains only on the concave families in S4. Its $\epsilon_{\mathrm{rel}}$ rises from 4.34\% in S1 to 20.46\% in S4, and its $\epsilon_{\mathrm{PDE}}$ from $6.43 \times 10^{-2}$ to $48.76 \times 10^{-2}$. For Client~1, trained only on Q and P in S4, $\epsilon_{\mathrm{rel}}$ rises from 4.36\% to 10.16\%. In contrast, the mean $\epsilon_{\mathrm{rel}}$ of Fed-PIGANO stays between 1.83\% and 2.04\% in all 12 cases and rises by at most 0.17 percentage points from S1 to S4. Its mean $\epsilon_{\mathrm{PDE}}$ stays between $3.34 \times 10^{-2}$ and $3.63 \times 10^{-2}$. Its accuracy ($\epsilon_{\mathrm{rel}}$) is also comparable to that of the pooled Centralized reference at every $\delta$ (Table~\ref{tab:results}). Through federation, each client benefits from the geometries held by the others without sharing its own. Therefore, its performance on unseen geometries is better than with local training alone and changes little as the family compositions of the clients diverge.

The results above are averages over the seven replications and over the 50 test geometries of each client. They do not show how the two metrics vary from one geometry to another. Each replication uses different test geometries, so such per-geometry values can be shown for only one replication at a time. Section~\ref{sec:results_geometry} examines one replication in detail.

\subsection{\emph{Prediction Accuracy on Individual Unseen Test Geometries}}
\label{sec:results_geometry}

\begin{figure}[htbp]
    \centering
    \begin{subfigure}[t]{0.25\textwidth}
        \centering
        \includegraphics[width=\linewidth]{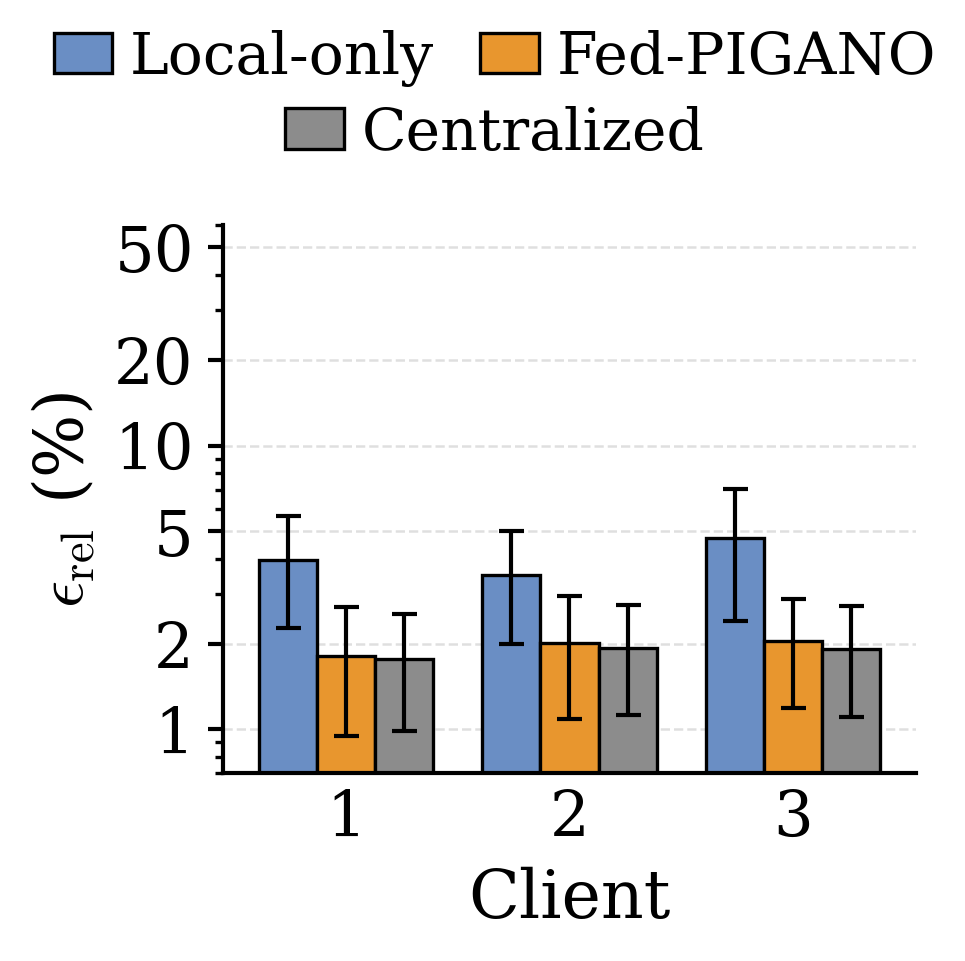}
        \caption{Scenario 1}
        \label{fig:rel_l2_S1}
    \end{subfigure}\hfill
    \begin{subfigure}[t]{0.25\textwidth}
        \centering
        \includegraphics[width=\linewidth]{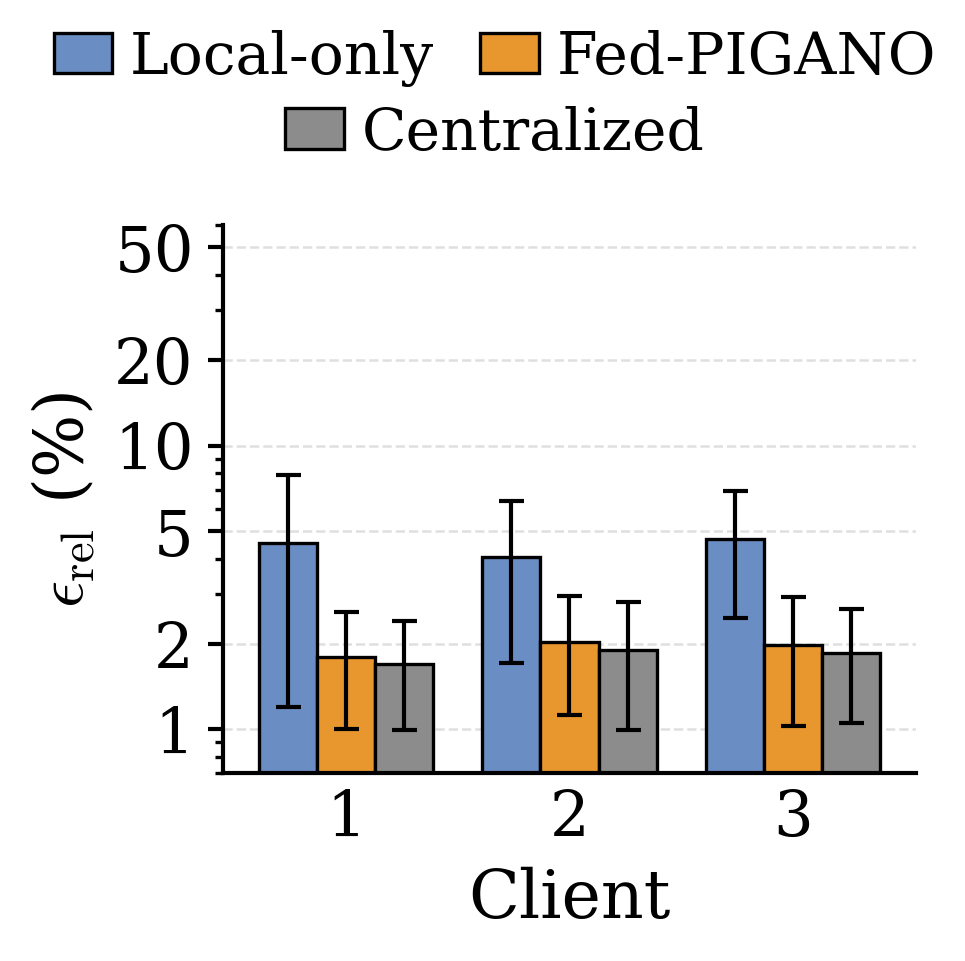}
        \caption{Scenario 2}
        \label{fig:rel_l2_S2}
    \end{subfigure}\hfill
    \begin{subfigure}[t]{0.25\textwidth}
        \centering
        \includegraphics[width=\linewidth]{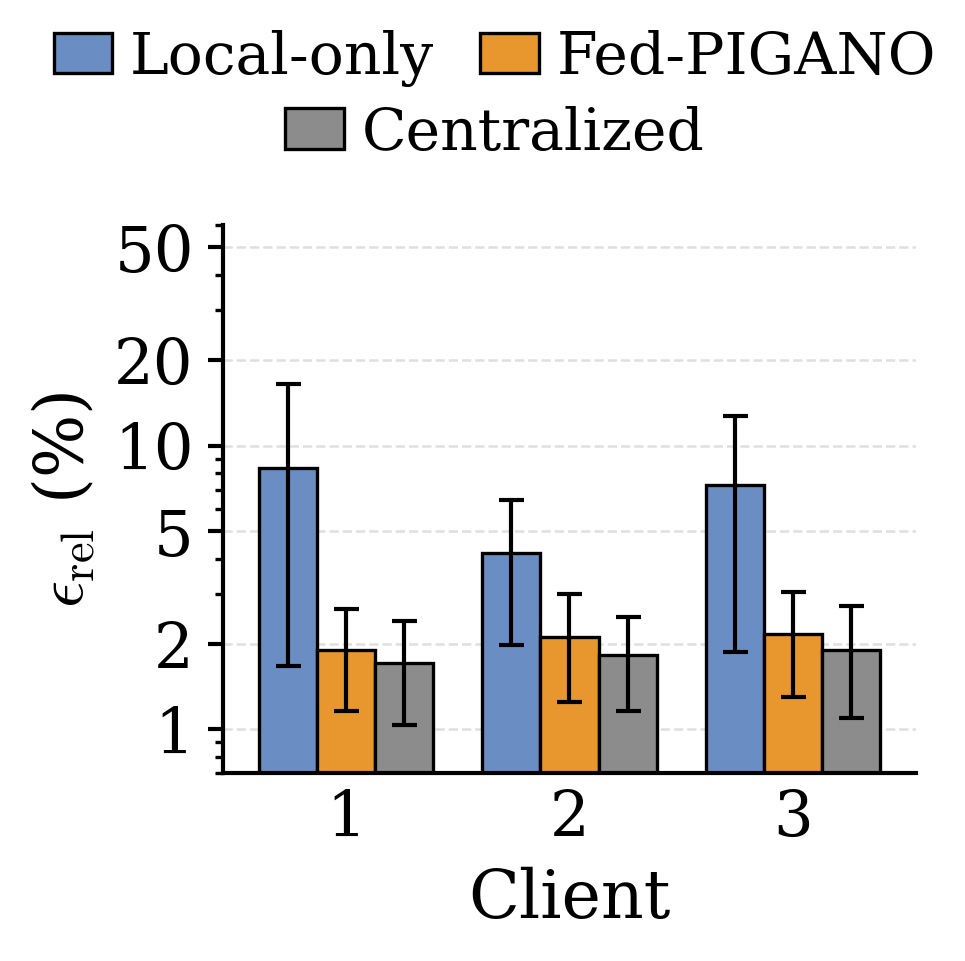}
        \caption{Scenario 3}
        \label{fig:rel_l2_S3}
    \end{subfigure}\hfill
    \begin{subfigure}[t]{0.25\textwidth}
        \centering
        \includegraphics[width=\linewidth]{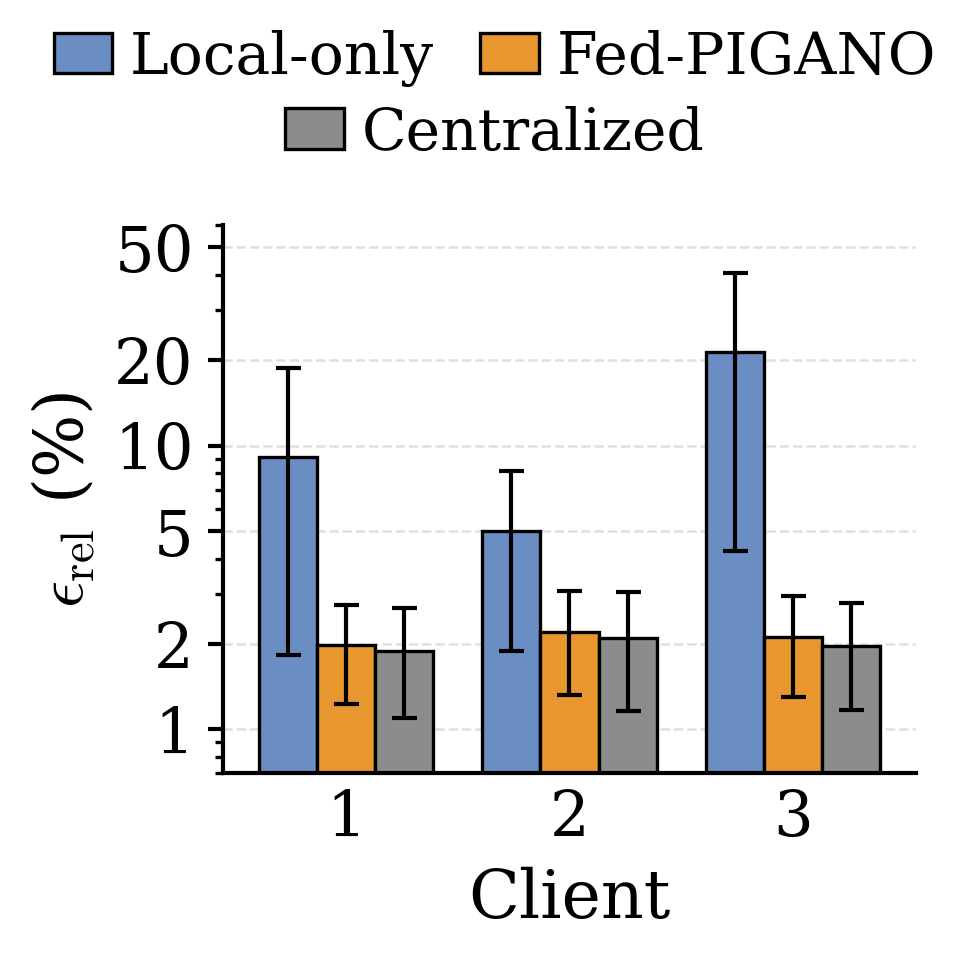}
        \caption{Scenario 4}
        \label{fig:rel_l2_S4}
    \end{subfigure}
    \caption{Mean $\epsilon_{\mathrm{rel}}$ across clients for (a) S1 to (d) S4. Error bars show $\pm 1$ standard deviation over the 50 test geometries. Lower error bars are cut at 20\% of the bar height where the standard deviation exceeds 80\% of the mean.}
    \label{fig:summary_rel_l2_all}
\end{figure}

\begin{figure}[htbp]
    \centering
    \begin{subfigure}[t]{0.25\textwidth}
        \centering
        \includegraphics[width=\linewidth]{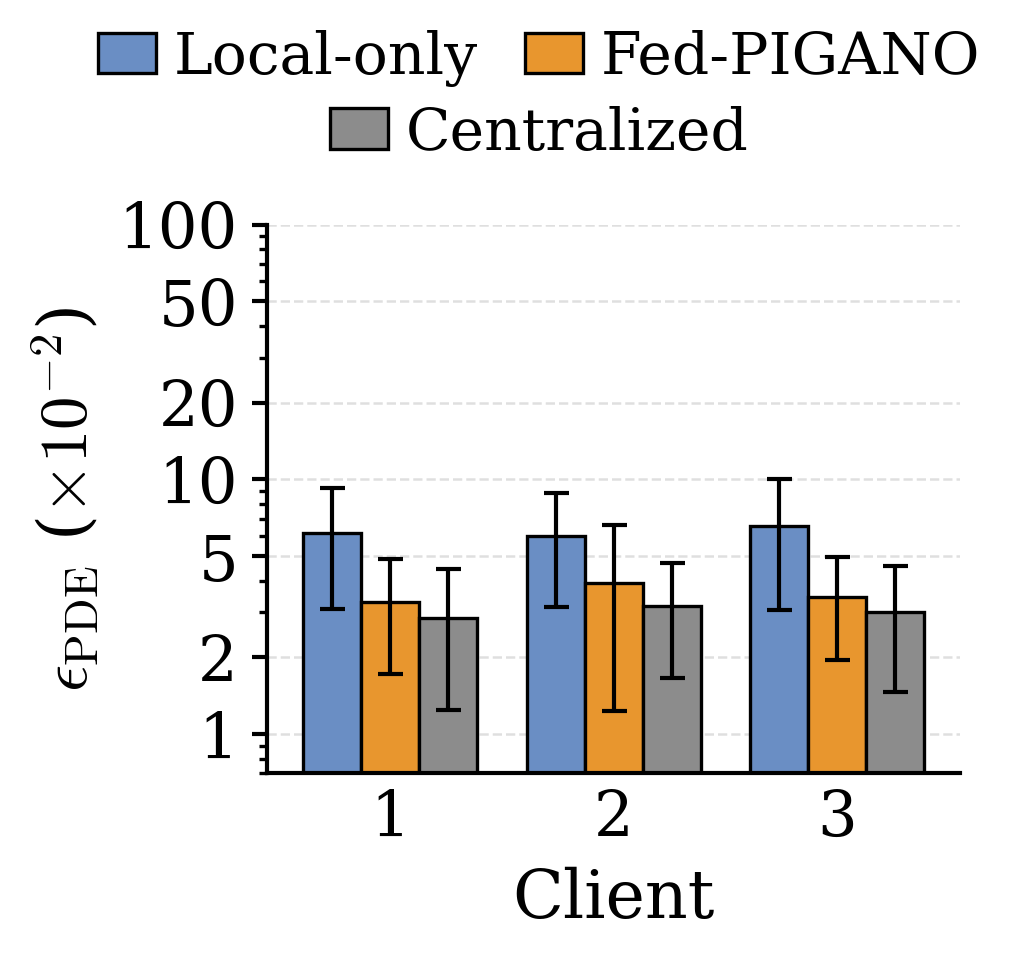}
        \caption{Scenario 1}
        \label{fig:pde_S1}
    \end{subfigure}\hfill
    \begin{subfigure}[t]{0.25\textwidth}
        \centering
        \includegraphics[width=\linewidth]{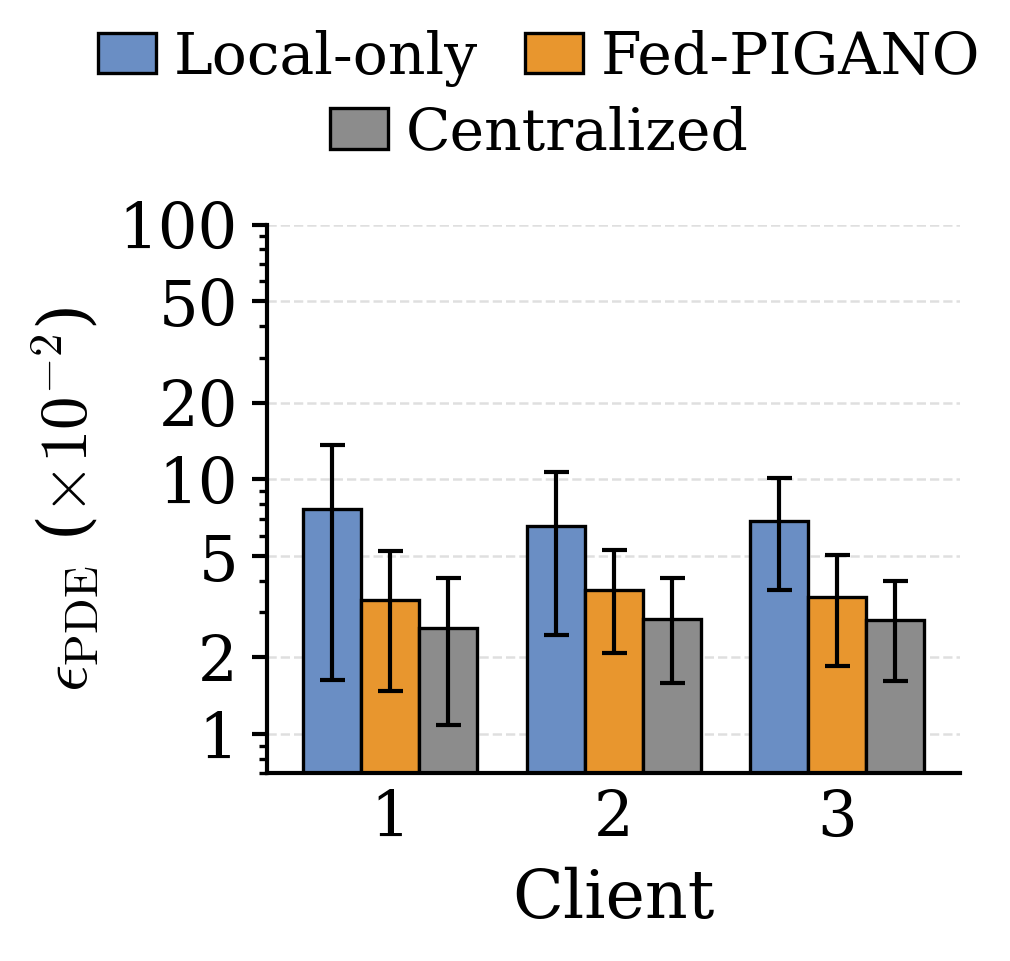}
        \caption{Scenario 2}
        \label{fig:pde_S2}
    \end{subfigure}\hfill
    \begin{subfigure}[t]{0.25\textwidth}
        \centering
        \includegraphics[width=\linewidth]{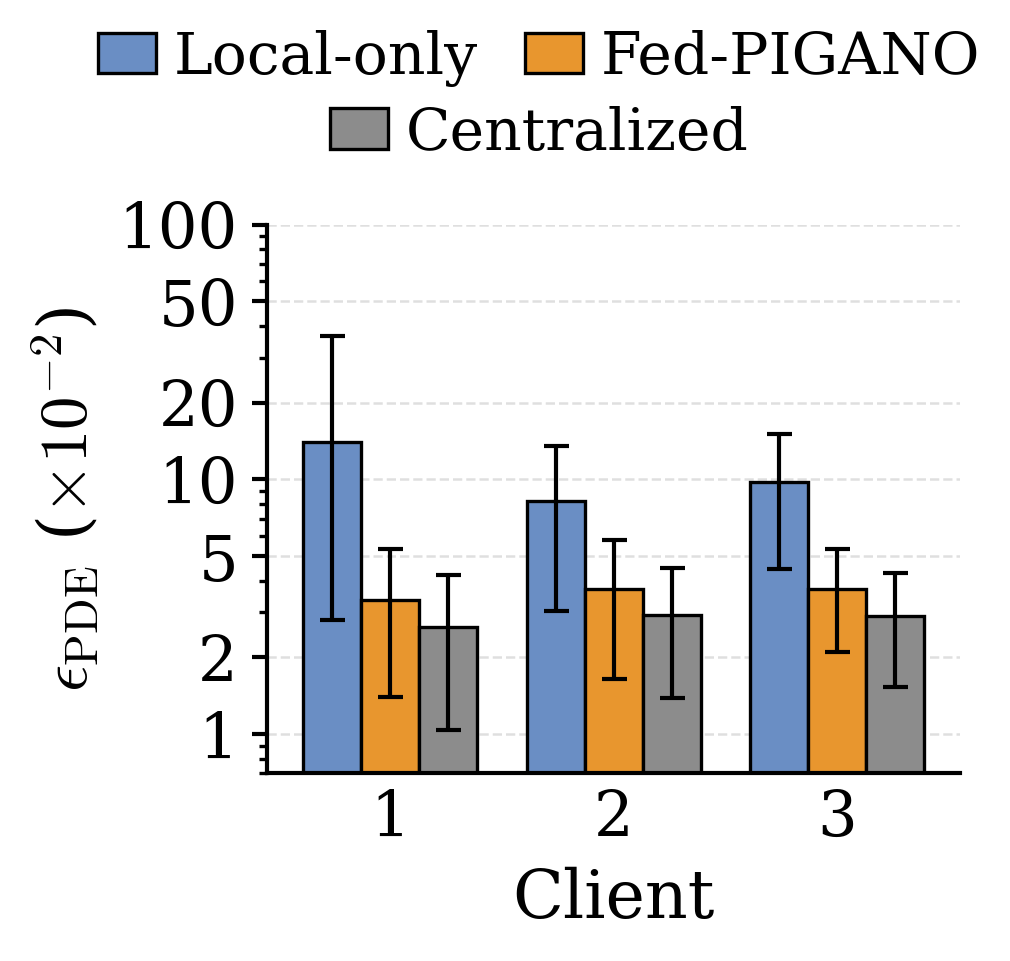}
        \caption{Scenario 3}
        \label{fig:pde_S3}
    \end{subfigure}\hfill
    \begin{subfigure}[t]{0.25\textwidth}
        \centering
        \includegraphics[width=\linewidth]{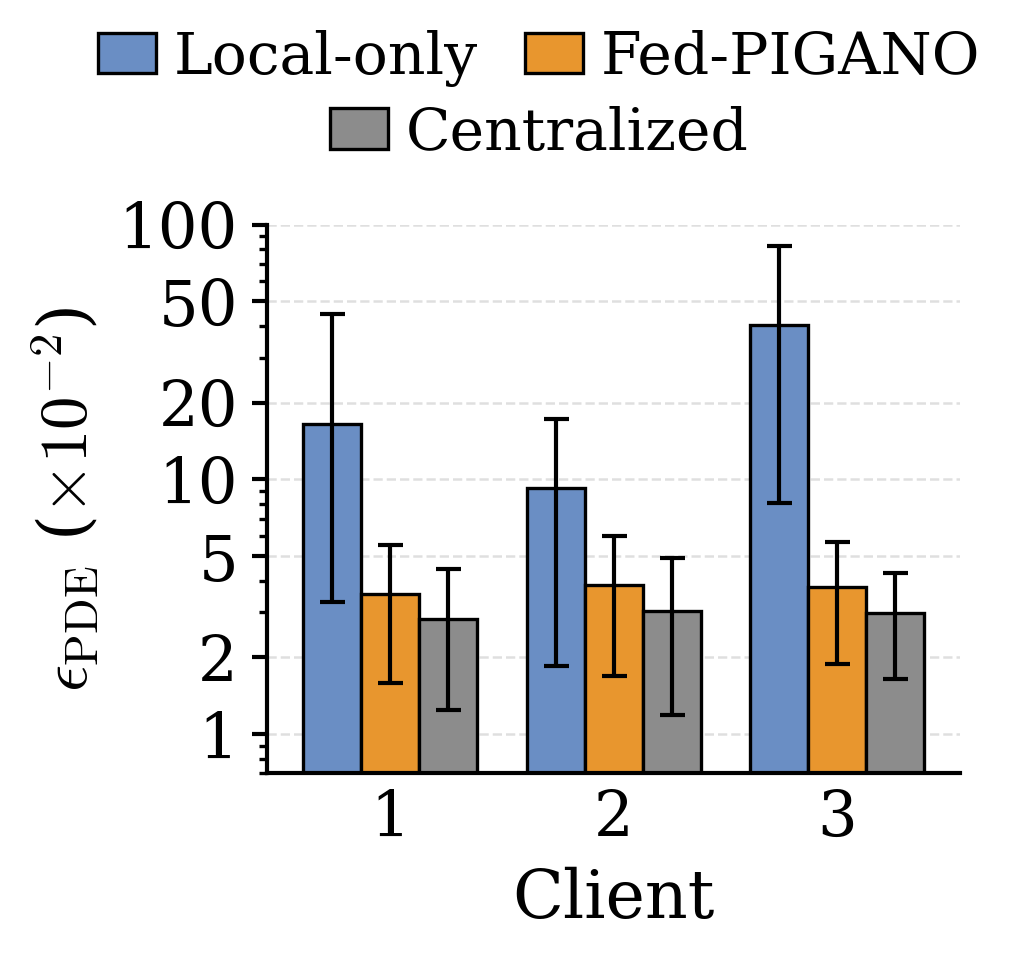}
        \caption{Scenario 4}
        \label{fig:pde_S4}
    \end{subfigure}
    \caption{Mean $\epsilon_{\mathrm{PDE}}$ across clients for (a) S1 to (d) S4. Error bars as in Fig.~\ref{fig:summary_rel_l2_all}.}
    \label{fig:summary_pde_all}
\end{figure}

We select one replication in which Fed-PIGANO performs slightly worse than its seven-replication average, in every scenario and for both metrics, and use it in this section and in Section~\ref{sec:results_fields}. The results of this replication are therefore conservative, since Fed-PIGANO performs better on average in the other six replications. For this replication, Figs.~\ref{fig:summary_rel_l2_all} and~\ref{fig:summary_pde_all} show the mean of each metric over the 50 test geometries of each client, with $\pm 1$ standard deviation over these geometries. As in Section~\ref{sec:results_mean}, Fed-PIGANO has a lower mean than Local-only for every client, in every scenario, and for both metrics. Its mean $\epsilon_{\mathrm{rel}}$ stays between 1.79\% and 2.20\% in all 12 cases, whereas that of Local-only lies between 3.51\% and 4.72\% in S1 and reaches 21.31\% for Client~3 in S4. The mean $\epsilon_{\mathrm{PDE}}$ shows the same contrast. It stays between $3.29 \times 10^{-2}$ and $3.93 \times 10^{-2}$ for Fed-PIGANO and reaches $40.39 \times 10^{-2}$ for Local-only. The spread over the test geometries differs in the same way. The standard deviation of the Local-only $\epsilon_{\mathrm{rel}}$ reaches 19.30\% for Client~3 in S4, while that of Fed-PIGANO stays between 0.75\% and 0.95\%. Centralized is included only as the pooled-training reference (Section~\ref{sec:local_baseline}).

\begin{figure}[!t]
    \centering
    \begin{subfigure}[t]{0.40\textwidth}
        \centering
        \includegraphics[width=\linewidth]{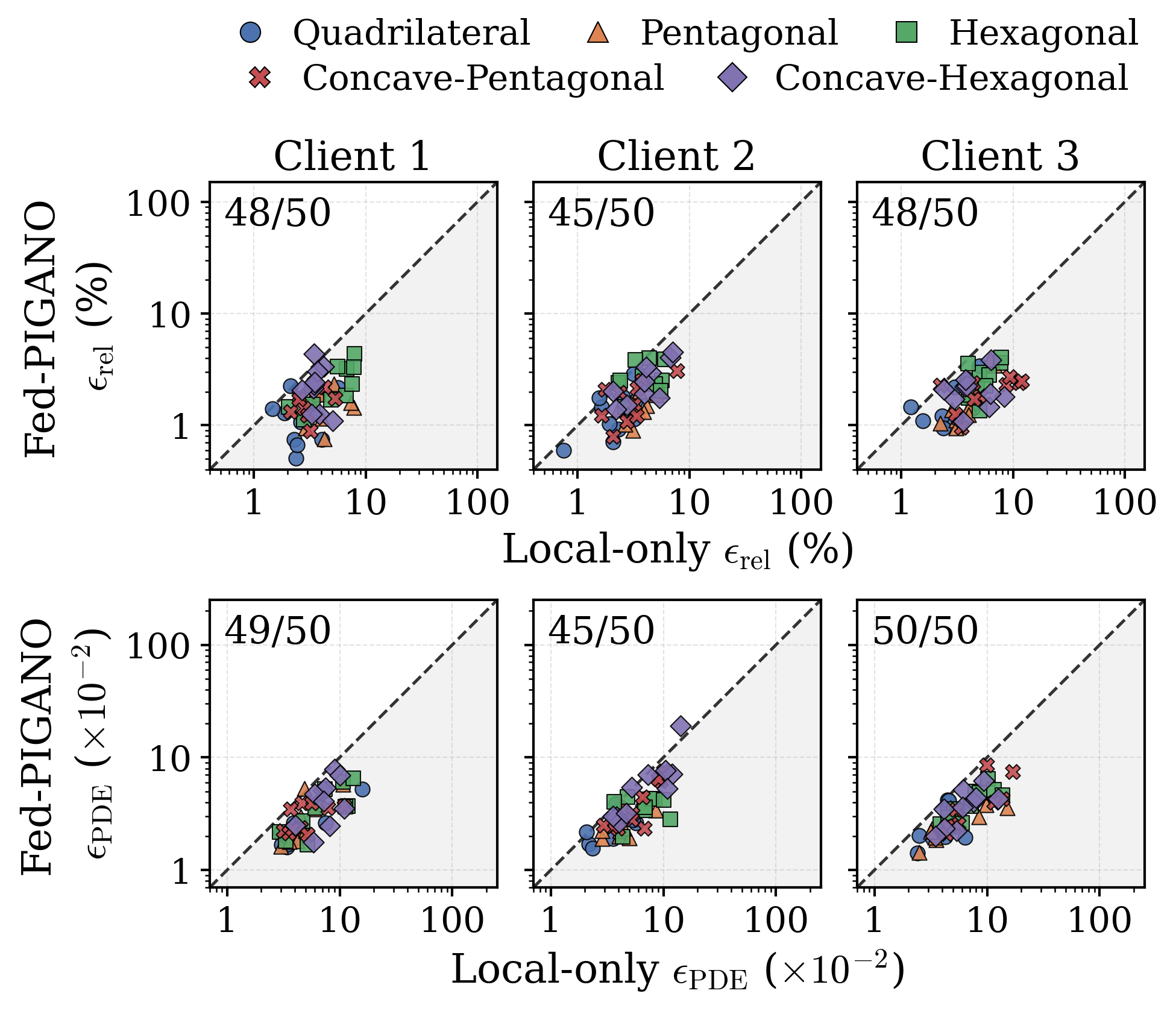}
        \caption{Scenario 1}
        \label{fig:parity_S1}
    \end{subfigure}
    \begin{subfigure}[t]{0.40\textwidth}
        \centering
        \includegraphics[width=\linewidth]{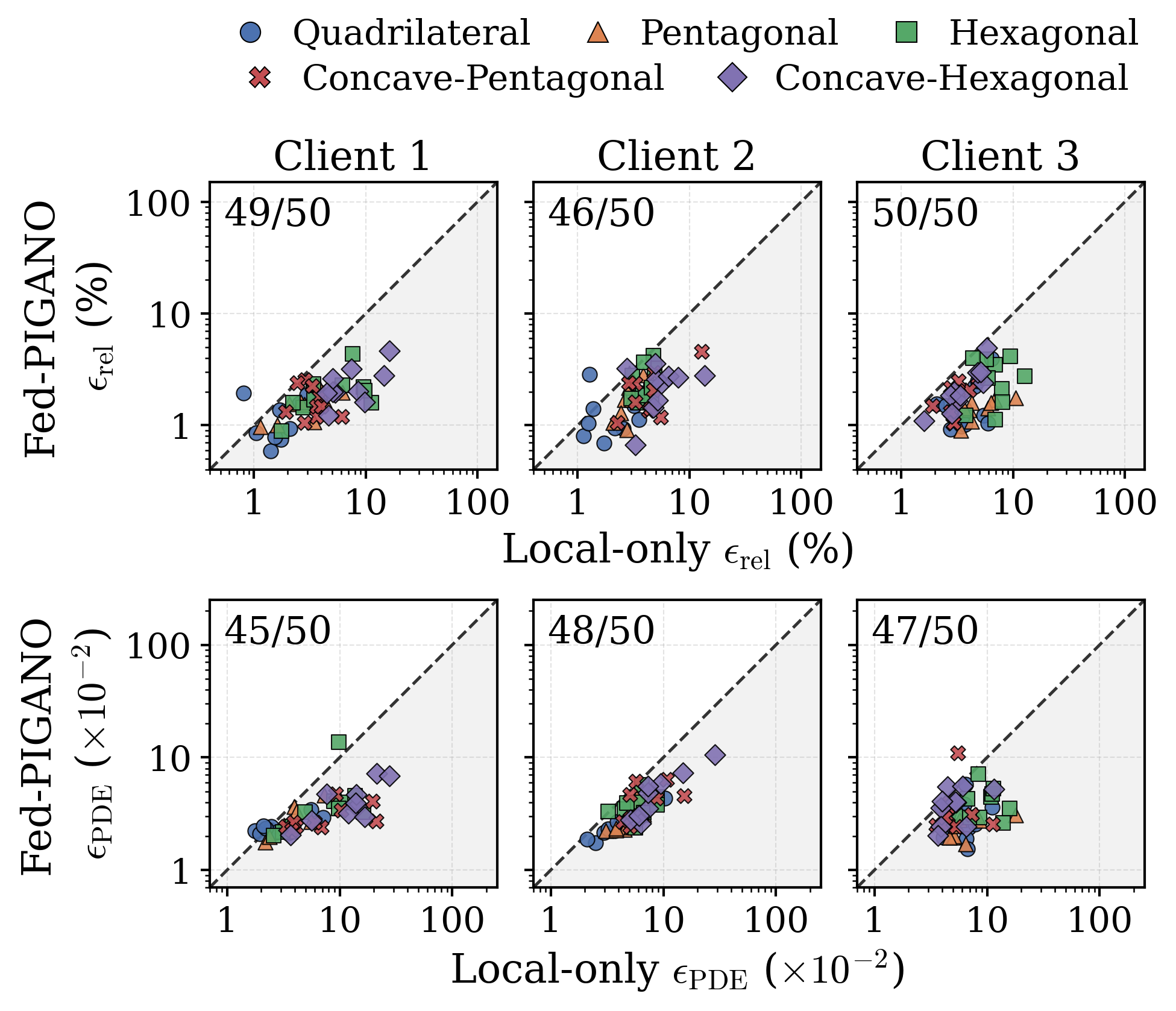}
        \caption{Scenario 2}
        \label{fig:parity_S2}
    \end{subfigure}

    \vspace{0.5em}
    \begin{subfigure}[t]{0.40\textwidth}
        \centering
        \includegraphics[width=\linewidth]{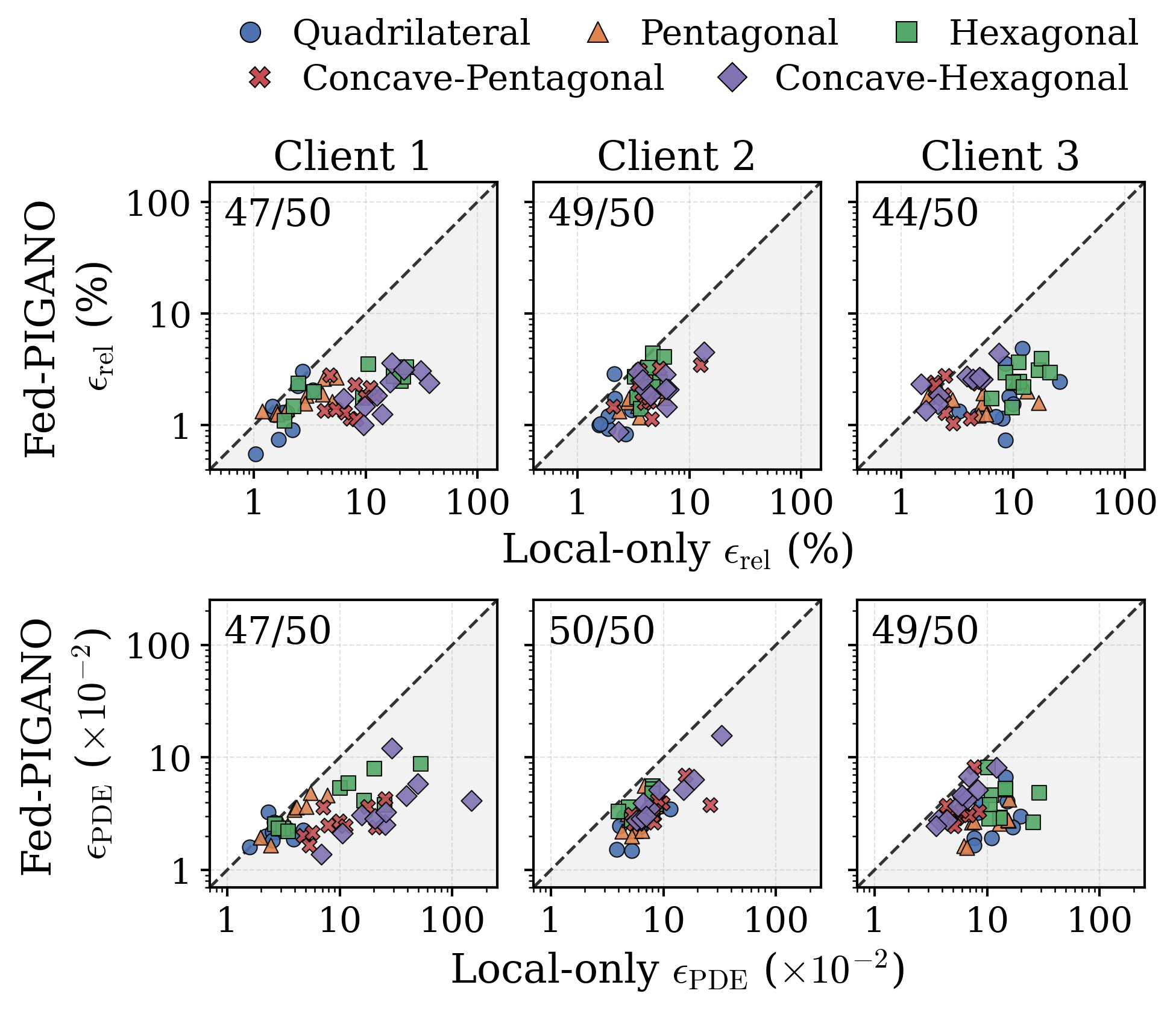}
        \caption{Scenario 3}
        \label{fig:parity_S3}
    \end{subfigure}
    \begin{subfigure}[t]{0.40\textwidth}
        \centering
        \includegraphics[width=\linewidth]{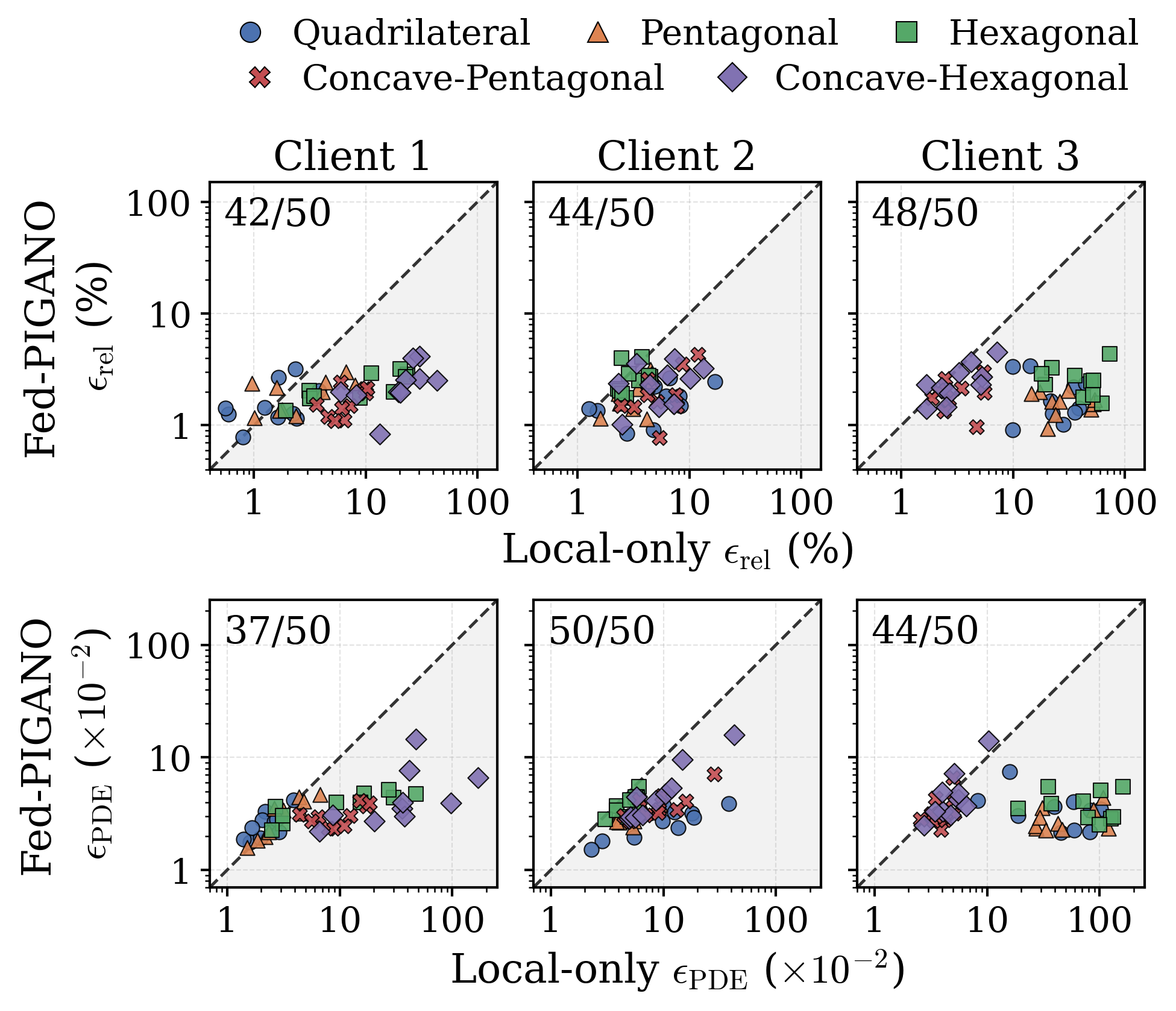}
        \caption{Scenario 4}
        \label{fig:parity_S4}
    \end{subfigure}
    \caption{Geometry-wise $\epsilon_{\mathrm{rel}}$ (top row of each subfigure) and $\epsilon_{\mathrm{PDE}}$ (bottom row) of Fed-PIGANO versus Local-only for (a) S1 to (d) S4. Markers below the identity line favor Fed-PIGANO.}
    \label{fig:parity_all}
\end{figure}

The statistics in Figs.~\ref{fig:summary_rel_l2_all} and~\ref{fig:summary_pde_all} still summarize all 50 test geometries of a client, so they do not show whether Fed-PIGANO is better on each geometry. Fig.~\ref{fig:parity_all} compares the two models geometry by geometry. Each marker is one test geometry, placed at its Local-only value on the horizontal axis and its Fed-PIGANO value on the vertical axis. Its shape and color give the geometry family. A marker below the dashed identity line, in the shaded region, marks a geometry on which Fed-PIGANO has the lower value, and the number in each panel counts these geometries. In every scenario--client case, Fed-PIGANO gives the lower $\epsilon_{\mathrm{rel}}$ on 42 to 50 of the 50 test geometries, and the lower $\epsilon_{\mathrm{PDE}}$ on 37 to 50. From S1 to S4, the markers shift to the right while staying at similar heights, as the Local-only values grow and the Fed-PIGANO values do not. The largest Local-only values occur on families that are absent from, or rare in, a client's training set. For Client~3 in S4, trained only on CP and CH, the 15 geometries with the largest Local-only $\epsilon_{\mathrm{rel}}$ all come from Q, P, and H. Therefore, Fed-PIGANO outperforms Local-only on most individual test geometries, not only on average, and its advantage is largest on the families that a client has not trained on. For a site, federation thus helps most on the new designs that its own data cover least.

\subsection{\emph{Predicted Thermal Fields on Sample Test Geometries}}
\label{sec:results_fields}

\begin{figure}[htbp]
    \centering
    \includegraphics[width=0.55\linewidth]{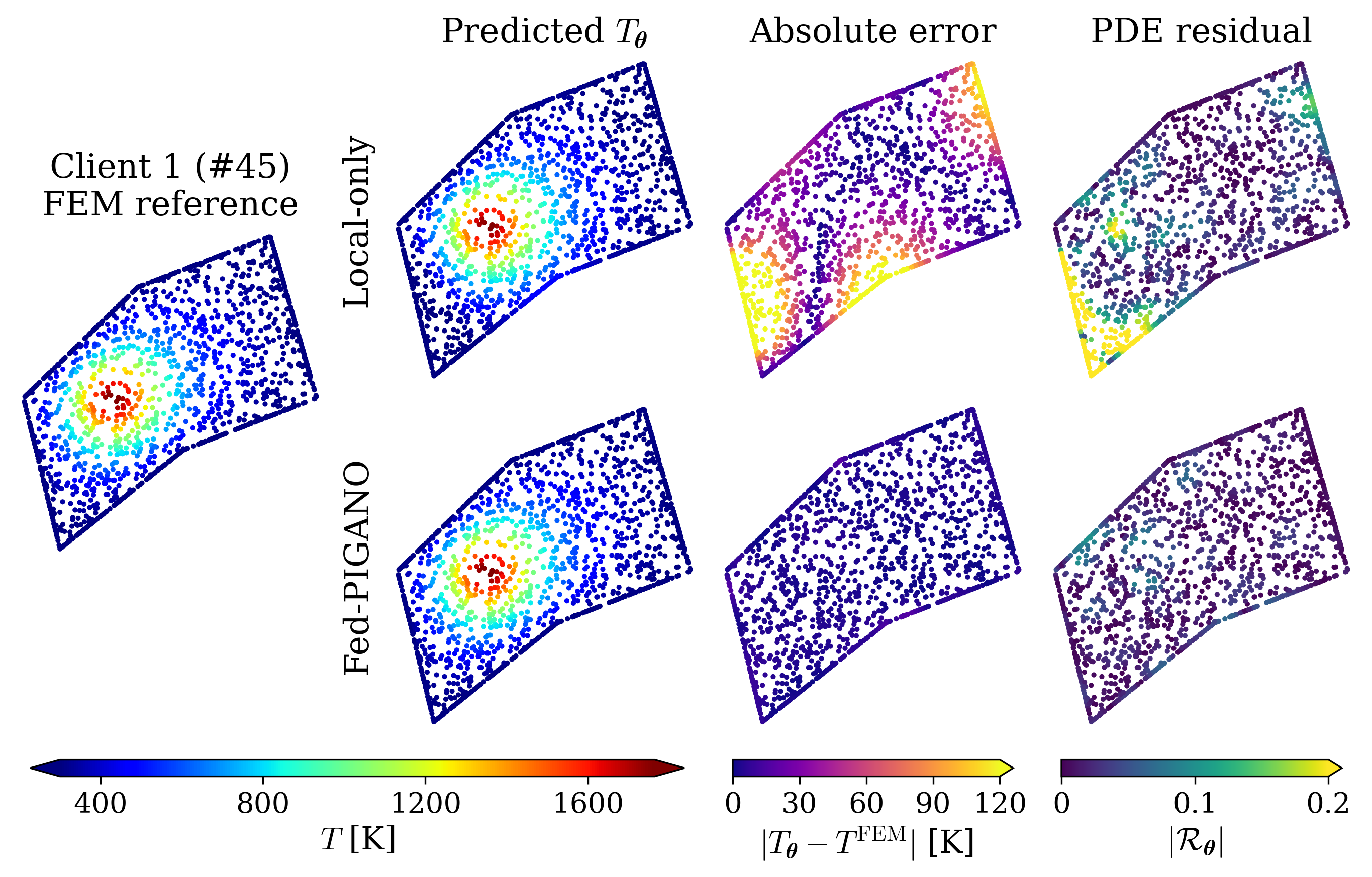}
    \caption{Thermal predictions for the unseen CH geometry \#45 at Client~1 in S4. The left panel shows the FEM reference. The rows show the predicted temperature, absolute error, and absolute PDE residual of each model. Error and residual color scales are shared across Figs.~\ref{fig:field_c1_S4}--\ref{fig:field_c3_S4}.}
    \label{fig:field_c1_S4}
\end{figure}

\begin{figure}[!t]
    \centering
    \includegraphics[width=0.55\linewidth]{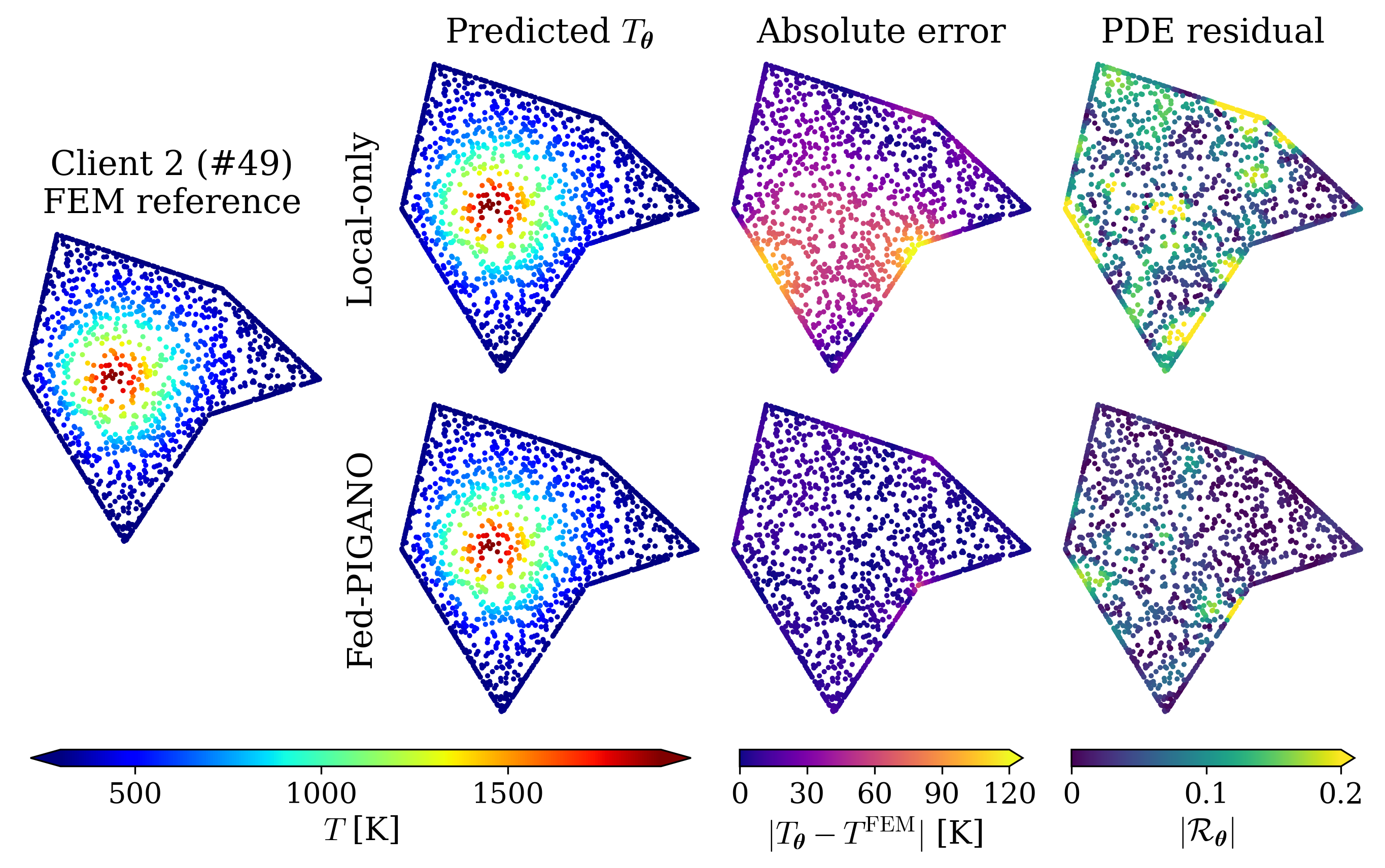}
    \caption{Thermal predictions for the unseen CH geometry \#49 at Client~2 in S4. Layout as in Fig.~\ref{fig:field_c1_S4}.}
    \label{fig:field_c2_S4}
\end{figure}

\begin{figure}[htbp]
    \centering
    \includegraphics[width=0.55\linewidth]{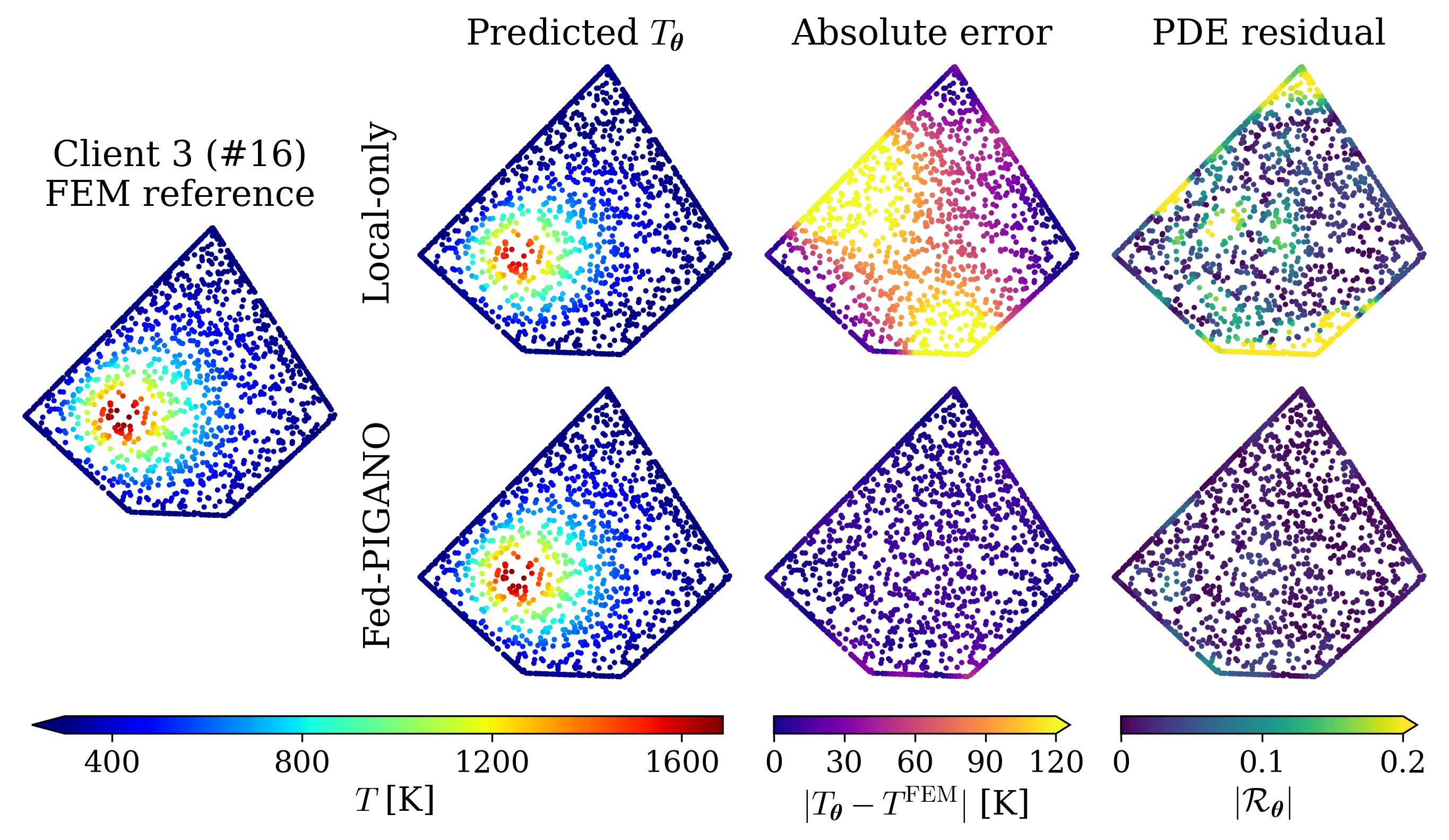}
    \caption{Thermal predictions for the unseen P geometry \#16 at Client~3 in S4. Layout as in Fig.~\ref{fig:field_c1_S4}.}
    \label{fig:field_c3_S4}
\end{figure}

This section shows predicted thermal fields from the replication of Section~\ref{sec:results_geometry}. We use S4, in which the family compositions of the clients differ most ($\delta = 0.768$). For each client, we select one unseen test geometry from a family that is absent from its training set (Table~\ref{tab:Fed_scenarios}). This mirrors a manufacturing site that receives a new geometric design not covered by its own training data. The selected geometries are CH for Clients~1 and~2, and P for Client~3. Figs.~\ref{fig:field_c1_S4}--\ref{fig:field_c3_S4} compare the FEM reference with the predicted temperature, absolute error, and absolute PDE residual of the client's Local-only model and of Fed-PIGANO. On these geometries, Local-only gives $\epsilon_{\mathrm{rel}}$ values of 13.30\%, 7.25\%, and 17.37\% at Clients~1, 2, and~3, and Fed-PIGANO gives 0.84\%, 1.55\%, and 1.98\%.

At the scale of the temperature range, the predictions of both models look similar to the FEM reference. However, the absolute error and PDE residual maps reveal a difference. On each geometry, the Fed-PIGANO error and residual stay small almost everywhere, whereas those of Local-only are much larger over wide regions. Fed-PIGANO is therefore more accurate than Local-only and satisfies the governing equation more closely. Figs.~\ref{fig:field_c1_S4}--\ref{fig:field_c3_S4} show only one geometry per client, but the per-geometry comparison in Fig.~\ref{fig:parity_all} confirms the same advantage of Fed-PIGANO on most test geometries. This advantage is largest on families absent from a client's training set. Thus, federation gives each client accurate predictions even on geometry families missing from its own data, without sharing any geometry.

\section{Conclusion and Future Work}
\label{sec:fed_conclusion_future_work}

We addressed thermal field prediction on unseen part geometries when the geometries are distributed across AM sites that cannot share them. We proposed Fed-PIGANO, which trains a physics-informed geometry-aware model at each site and aggregates its geometry encoder and prediction network across sites. We evaluated it on 2D metal AM geometries from five families, held by three sites under four scenarios of increasing heterogeneity, over seven replications. Fed-PIGANO was more accurate than training at each site alone (Local-only) in all 12 scenario--site cases and in every replication. Even with identical family compositions, the mean relative error of Local-only was about 2.3 times that of Fed-PIGANO. As the compositions diverged, it rose up to 20.46\%, while that of Fed-PIGANO stayed between 1.83\% and 2.04\%. The accuracy of Fed-PIGANO was also comparable to that of a centralized model trained on the pooled geometries. Future work will extend the framework to 3D geometries and transient AM thermal problems with moving heat sources, investigate coupling with microstructure and property models, and explore secure aggregation and differential privacy to further protect proprietary part designs.

\if0\blind{
\section*{Acknowledgements}
    During the preparation of this manuscript, the authors used large language models (LLMs) for language editing of selected portions of the text. The authors reviewed and edited all LLM-assisted content and take full responsibility for the final manuscript.
} \fi

\bibliographystyle{apalike}
\spacingset{1}
\bibliography{IISE-Trans}
	
\end{document}